\documentclass[10pt]{article}
\usepackage{arxiv}

\usepackage[utf8]{inputenc} 
\usepackage[T1]{fontenc}    
\usepackage{hyperref}       
\usepackage{url}            
\usepackage{booktabs}       
\usepackage{amsfonts}       
\usepackage{nicefrac}       
\usepackage{microtype}      
\usepackage{lipsum}

\usepackage{natbib}
\setcitestyle{numbers,sort&compress,comma,open={[},close={]}}

\usepackage{subcaption}
\usepackage[utf8]{inputenc}
\usepackage[export]{adjustbox}
\usepackage{wrapfig}
\usepackage{xcolor}
\usepackage{rotating}

\usepackage{caption}
\usepackage{graphicx}
\graphicspath{{images/}}

\usepackage{amsmath}
\usepackage{array}

\usepackage{stfloats}

\usepackage{url}

\usepackage{tikz}
\usetikzlibrary{shapes.geometric, arrows}
\usetikzlibrary{positioning}
\usetikzlibrary{decorations.pathreplacing}
\usepackage{varwidth}

\usepackage{multirow}

\usepackage[inline]{enumitem} 
\usepackage{rotating} 
\usepackage[acronym,nomain,nonumberlist]{glossaries}

\usepackage[title,page]{appendix}

\makeglossaries
\newacronym{dl}{DL}{deep learning}
\newacronym{eeg}{EEG}{electroencephalography}
\newacronym{meg}{MEG}{magnetoencephalography}
\newacronym{emg}{EMG}{electromyography}
\newacronym{eog}{EOG}{electroculography}
\newacronym{ecog}{ECoG}{electrocorticography}
\newacronym{snr}{SNR}{signal-to-noise ratio}
\newacronym{bci}{BCI}{brain-computer interface}
\newacronym{erp}{ERP}{event-related potential}
\newacronym{rsvp}{RSVP}{rapid serial visual presentation}

\newacronym{dnn}{DNN}{deep neural network}
\newacronym{cnn}{CNN}{convolutional neural network}
\newacronym{rnn}{RNN}{recurrent neural network}
\newacronym{lstm}{LSTM}{long short-term memory}
\newacronym{dbn}{DBN}{deep belief network}
\newacronym{rbm}{RBM}{restricted Boltzmann machine}
\newacronym{gan}{GAN}{generative adversarial network}
\newacronym{ae}{AE}{autoencoder}
\newacronym{sdae}{SDAE}{stacked denoising autoencoder}
\newacronym{fc}{FC}{fully-connected}

\newacronym{nlp}{NLP}{natural language processing}
\newacronym{cv}{CV}{computer vision}
\newacronym{sgd}{SGD}{stochastic gradient descent}
\newacronym{rocauc}{ROC AUC}{area under the receiver operating curve}

\newacronym{svm}{SVM}{support vector machine}
\newacronym{ica}{ICA}{independent component analysis}
\newacronym{stft}{STFT}{short-time Fourier transform}
\newacronym{psd}{PSD}{power spectral density}
\newacronym{cca}{CCA}{canonical correlation analysis}

\begin{document}

\title{Deep learning-based electroencephalography analysis: a systematic review}

\author{
  Yannick Roy\thanks{The first two authors contributed equally to this work.} \\
  Faubert Lab \\
  Universit\'e de Montr\'eal\\
  Montr\'eal, Canada \\
  \texttt{yannick.roy@umontreal.ca} \\
  \And
  Hubert Banville\footnotemark[1] \\
  Inria\\
  Universit\'e Paris-Saclay\\
  Paris, France \& \\
  InteraXon Inc.\\
  Toronto, Canada\\
  \And
  Isabela Albuquerque \\
  MuSAE Lab \\
  INRS-EMT\\
  Universit\'e du Qu\'ebec\\
  Montr\'eal, Canada \\
  \And
  Alexandre Gramfort \\
  Inria\\
  Universit\'e Paris-Saclay\\
  Paris, France \\
  \And
  Tiago H. Falk \\
  MuSAE Lab \\
  INRS-EMT\\
  Universit\'e du Qu\'ebec\\
  Montr\'eal, Canada \\
  \And
  Jocelyn Faubert \\
  Faubert Lab \\
  Universit\'e de Montr\'eal\\
  Montr\'eal, Canada \\
}

\maketitle

\begin{abstract}

\textbf{Context}. Electroencephalography (EEG) is a complex signal and can require several years of training, as well as advanced signal processing and feature extraction methodologies to be correctly interpreted. 
Recently, deep learning (DL) has shown great promise in helping make sense of EEG signals due to its capacity to learn good feature representations from raw data. 
Whether DL truly presents advantages as compared to more traditional EEG processing approaches, however, remains an open question.

\textbf{Objective}. In this work, we review 156 papers that apply DL to EEG, published between January 2010 and July 2018, and spanning different application domains such as epilepsy, sleep, brain-computer interfacing, and cognitive and affective monitoring.
We extract trends and highlight interesting approaches from this large body of literature in order to inform future research and formulate recommendations.
 
\textbf{Methods}. Major databases spanning the fields of science and engineering were queried to identify relevant studies published in scientific journals, conferences, and electronic preprint repositories. 
Various data items were extracted for each study pertaining to 1) the data, 2) the preprocessing methodology, 3) the DL design choices, 4) the results, and 5) the reproducibility of the experiments.
These items were then analyzed one by one to uncover trends.

\textbf{Results}. 
Our analysis reveals that the amount of EEG data used across studies varies from less than ten minutes to thousands of hours, while the number of samples seen during training by a network varies from a few dozens to several millions, depending on how epochs are extracted. Interestingly, we saw that more than half the studies used publicly available data and that there has also been a clear shift from intra-subject to inter-subject approaches over the last few years.
About $40\%$ of the studies used convolutional neural networks (CNNs), while $14\%$ used recurrent neural networks (RNNs), most often with a total of 3 to 10 layers.
Moreover, almost one-half of the studies trained their models on raw or preprocessed EEG time series.
Finally, the median gain in accuracy of DL approaches over traditional baselines was $5.4\%$ across all relevant studies.
More importantly, however, we noticed studies often suffer from poor reproducibility: a majority of papers would be hard or impossible to reproduce given the unavailability of their data and code.

\textbf{Significance}. To help the community progress and share work more effectively, we provide a list of recommendations for future studies. We also make our summary table of DL and EEG papers available and invite authors of published work to contribute to it directly.
\end{abstract}




\tikzset{%
  every neuron/.style={
    circle,
    draw,
    minimum size=0.8cm
  },
  neuron missing/.style={
    draw=none, 
    scale=1,
    execute at begin node=\color{black}$\vdots$
  },
}



\section{Introduction} \label{sec:introduction}

\subsection{Measuring brain activity with EEG}

\Gls{eeg}, the measure of the electrical fields produced by the active brain, is a neuroimaging technique widely used inside and outside the clinical domain.
Specifically, \gls{eeg} picks up the electric potential differences, on the order of tens of $\mu V$, that reach the scalp when tiny excitatory post-synaptic potentials produced by pyramidal neurons in the cortical layers of the brain sum together.
The potentials measured therefore reflect neuronal activity and can be used to study a wide array of brain processes.

Thanks to the great speed at which electric fields propagate, \gls{eeg} has an excellent temporal resolution: events occurring at millisecond timescales can typically be captured.
However, \gls{eeg} suffers from low spatial resolution, as the electric fields generated by the brain are smeared by the tissues, such as the skull, situated between the sources and the sensors.
As a result, \gls{eeg} channels are often highly correlated spatially.
The source localization problem, or inverse problem, is an active area of research in which algorithms are developed to reconstruct brain sources given \gls{eeg} recordings \cite{he2018electro}.

There are many applications for \gls{eeg}.
For example, in clinical settings, \gls{eeg} is often used to study sleep patterns \cite{aboalayon2016sleep} or epilepsy \cite{acharya2013automated}.
Various conditions have also been linked to changes in electrical brain activity, and can therefore be monitored to various extents using \gls{eeg}. These include attention deficit hyperactivity disorder (ADHD) \cite{arns2013decade}, disorders of consciousness \cite{giacino2014disorders, engemann2018robust}, depth of anaesthesia \cite{hagihira2015changes}, etc.
\gls{eeg} is also widely used in neuroscience and psychology research, as it is an excellent tool for studying the brain and its functioning.
Applications such as cognitive and affective monitoring are very promising as they could allow unbiased measures of, for example, an individual's level of fatigue, mental workload, \cite{berka2007eeg, zander2011towards}, mood, or emotions \cite{al2017review}.
Finally, \gls{eeg} is widely used in \glspl{bci} - communication channels that bypass the natural output pathways of the brain - to allow brain activity to be directly translated into directives that affect the user's environment \cite{lotte2015electroencephalography}.

\subsection{Current challenges in EEG processing}

Although \gls{eeg} has proven to be a critical tool in many domains, it still suffers from a few limitations that hinder its effective analysis or processing.
First, \gls{eeg} has a low \gls{snr} \cite{Bigdely-Shamlo-etal:16,jas-etal:17}, as the brain activity measured is often buried under multiple sources of environmental, physiological and activity-specific noise of similar or greater amplitude called ``artifacts''.
Various filtering and noise reduction techniques have to be used therefore to minimize the impact of these noise sources and extract true brain activity from the recorded signals.

\gls{eeg} is also a non-stationary signal \cite{Cole302000,gramfortetal2013}, that is its statistics vary across time.
As a result, a classifier trained on a temporally-limited amount of user data might generalize poorly to data recorded at a different time on the same individual.
This is an important challenge for real-life applications of \gls{eeg}, which often need to work with limited amounts of data.

Finally, high inter-subject variability also limits the usefulness of \gls{eeg} applications.
This phenomenon arises due to physiological differences between individuals, which vary in magnitude but can severely affect the performance of models that are meant to generalize across subjects \cite{clerc2016brain}.
Since the ability to generalize from a first set of individuals to a second, unseen set is key to many practical applications of \gls{eeg}, a lot of effort is being put into developing methods that can handle inter-subject variability.

To solve some of the above-mentioned problems, processing pipelines with domain-specific approaches are often used.
A significant amount of research has been put into developing processing pipelines to clean, extract relevant features, and classify \gls{eeg} data.
State-of-the-art techniques, such as Riemannian geometry-based classifiers and adaptive classifiers \cite{lotte2018review}, can handle these problems with varying levels of success.

Additionally, a wide variety of tasks would benefit from a higher level of automated processing.
For example, sleep scoring, the process of annotating sleep recordings by categorizing windows of a few seconds into sleep stages, currently requires a lot of time, being done manually by trained technicians.
More sophisticated automated \gls{eeg} processing could make this process much faster and more flexible.
Similarly, real-time detection or prediction of the onset of an epileptic seizure would be very beneficial to epileptic individuals, but also requires automated \gls{eeg} processing.
For each of these applications, most common implementations require domain-specific processing pipelines, which further reduces the flexibility and generalization capability of current \gls{eeg}-based technologies.

\subsection{Improving EEG processing with deep learning}

To overcome the challenges described above, new approaches are required to improve the processing of \gls{eeg} towards better generalization capabilities and more flexible applications.
In this context, \gls{dl} \cite{lecun2015deep} could significantly simplify processing pipelines by allowing automatic end-to-end learning of preprocessing, feature extraction and classification modules, while also reaching competitive performance on the target task.
Indeed, in the last few years, \gls{dl} architectures have been very successful in processing complex data such as images, text and audio signals~\cite{lecun2015deep}, leading to state-of-the-art performance on multiple public benchmarks - such as the Large Scale Visual Recognition challenge \cite{deng2012ilsvrc} - and an ever-increasing role in industrial applications.

\Gls{dl}, a subfield of machine learning, studies computational models that learn hierarchical representations of input data through successive non-linear transformations \cite{lecun2015deep}.
\Glspl{dnn}, inspired by earlier models such as the perceptron \cite{rosenblatt1958perceptron}, are models where: 1) stacked layers of artificial ``neurons'' each apply a linear transformation to the data they receive and 2) the result of each layer's linear transformation is fed through a non-linear activation function.
Importantly, the parameters of these transformations are learned by directly minimizing a cost function.
Although the term ``deep'' implies the inclusion of many layers, there is no consensus on how to measure depth in a neural network and therefore on what really constitutes a deep network and what does not \cite{goodfellow2016deep}.

Fig.~\ref{fig:intro_dleeg} presents an overview of how \gls{eeg} data (and similar multivariate time series) can be formatted to be fed into a \gls{dl} model, along with some important terminology (see Section~\ref{ss:terminology}), as well as an illustration of a generic neural network architecture.
Usually, when $c$ channels are available and a window has length $l$ samples, the input of a neural network for \gls{eeg} processing consists of a multidimensional array $X_i \in \mathbb{R}^{c \times l}$ containing the $l$ samples corresponding to a window for all channels. This multidimensional array can be used as an example for training a neural network, as shown in Fig.~\ref{fig:dl}. 
Variations of this end-to-end formulation can be imagined where the window $X_i$ is first passed through a preprocessing and feature extraction pipeline (e.g., time-frequency transform), yielding an example $X_i'$ which is then used as input to the neural network instead.

Different types of layers are used as building blocks in neural networks. 
Most commonly, those are \gls{fc}, convolutional or recurrent layers.
We refer to models using these types of layers as \gls{fc} networks, \glspl{cnn} \cite{lecun1989backpropagation} and \glspl{rnn} \cite{rumelhart1986learning}, respectively.
Here, we provide a quick overview of the main architectures and types of models. 
The interested reader is referred to the relevant literature for more in-depth descriptions of \gls{dl} methodology \cite{lecun2015deep, goodfellow2016deep, schmidhuber2015deep}.

\Gls{fc} layers are composed of fully-connected neurons, i.e., where each neuron receives as input the activations of every single neuron of the preceding layer.
Convolutional layers, on the other hand, impose a particular structure where neurons in a given layer only see a subset of the activations of the preceding one.
This structure, akin to convolutions in signal or image processing from which it gets its name, encourages the model to learn invariant representations of the data. 
This property stems from another fundamental characteristic of convolutional layers, which is that parameters are shared across different neurons - this can be interpreted as if there were filters looking for the same information across patches of the input. 
In addition, pooling layers can be introduced, such that the representations learned by the model become invariant to slight translations of the input. 
This is often a desirable property: for instance, in an object recognition task, translating the content of an image should not affect the prediction of the model.
Imposing these kinds of priors thus works exceptionally well on data with spatial structure.
In contrast to convolutional layers, recurrent layers impose a structure by which, in its most basic form, a layer receives as input both the preceding layer's current activations and its own activations from a previous time step.
Models composed of recurrent layers are thus encouraged to make use of the temporal structure of data and have shown high performance in \gls{nlp} tasks \cite{zhou2015end, yogatama2017generative}.

Additionally, outside of purely supervised tasks, other architectures and learning strategies can be built to train models when no labels are available.
For example, \glspl{ae} learn a representation of the input data by trying to reproduce their input given some constraints, such as sparsity or the introduction of artificial noise \cite{goodfellow2016deep}.
\Glspl{gan} \cite{goodfellow2014generative} are trained by opposing a generator (G), that tries to generate fake examples from an unknown distribution of interest, to a discriminator (D), that tries to identify whether the input it receives has been artificially generated by G or is an example from the unknown distribution of interest. This dynamic can be compared to the one between a thief (G) making fake money and the police (D) trying to distinguish fake money from real money. Both agents push one another to get better, up to a point where the fake money looks exactly like real money.
The training of G and D can thus be interpreted as a two-player zero-sum minimax game.
When equilibrium is reached, the probability distribution approximated by G converges to the real data distribution \cite{goodfellow2014generative}.

Overall, there are multiple ways in which \gls{dl} improve and extend existing \gls{eeg} processing methods.
First, the hierarchical nature of \glspl{dnn} means features could potentially be learned on raw or minimally preprocessed data, reducing the need for domain-specific processing and feature extraction pipelines.
Features learned through a \gls{dnn} might also be more effective or expressive than the ones engineered by humans.
Second, as is the case in the multiple domains where \gls{dl} has surpassed the previous state-of-the-art, it has the potential to produce higher levels of performance on different analysis tasks.
Third, \gls{dl} facilitates the development of tasks that are less often attempted on \gls{eeg} data such as generative modelling 
\cite{goodfellow2016nips} 
and transfer learning \cite{pan2010survey}.
Indeed, generative models can be leveraged to learn intermediate representations or for data augmentation \cite{goodfellow2016nips}.
In transfer learning, the model parameters can also be transferred from one subject to another or from task A to task B. This might drastically widen or change the applicability of several \gls{eeg}-based technologies.

On the other hand, there are various reasons why \gls{dl} might not be optimal for \gls{eeg} processing and that may justify the skepticism of some of the \gls{eeg} community.
First and foremost, the datasets typically available in \gls{eeg} research contain far fewer examples than what has led to the current state-of-the-art in \gls{dl}-heavy domains such as \gls{cv} and \gls{nlp}.
Data collection being relatively expensive and data accessibility often being hindered by privacy concerns - especially with clinical data - openly available datasets of similar sizes are not common.
Some initiatives have tried to tackle this problem though \cite{harati2014tuh}.
Second, the peculiarities of \gls{eeg}, such as its low \gls{snr}, make \gls{eeg} data very different from other types of data (e.g, images, text and speech) for which \gls{dl} has been most successful.
Therefore, the architectures and practices that are currently used in \gls{dl} might not be readily applicable to \gls{eeg} processing.

\subsection{Terminology used in this review}
\label{ss:terminology}

\begin{figure}
    \begin{subfigure}{\textwidth}
    \centering
    \includegraphics[width=0.75\textwidth]{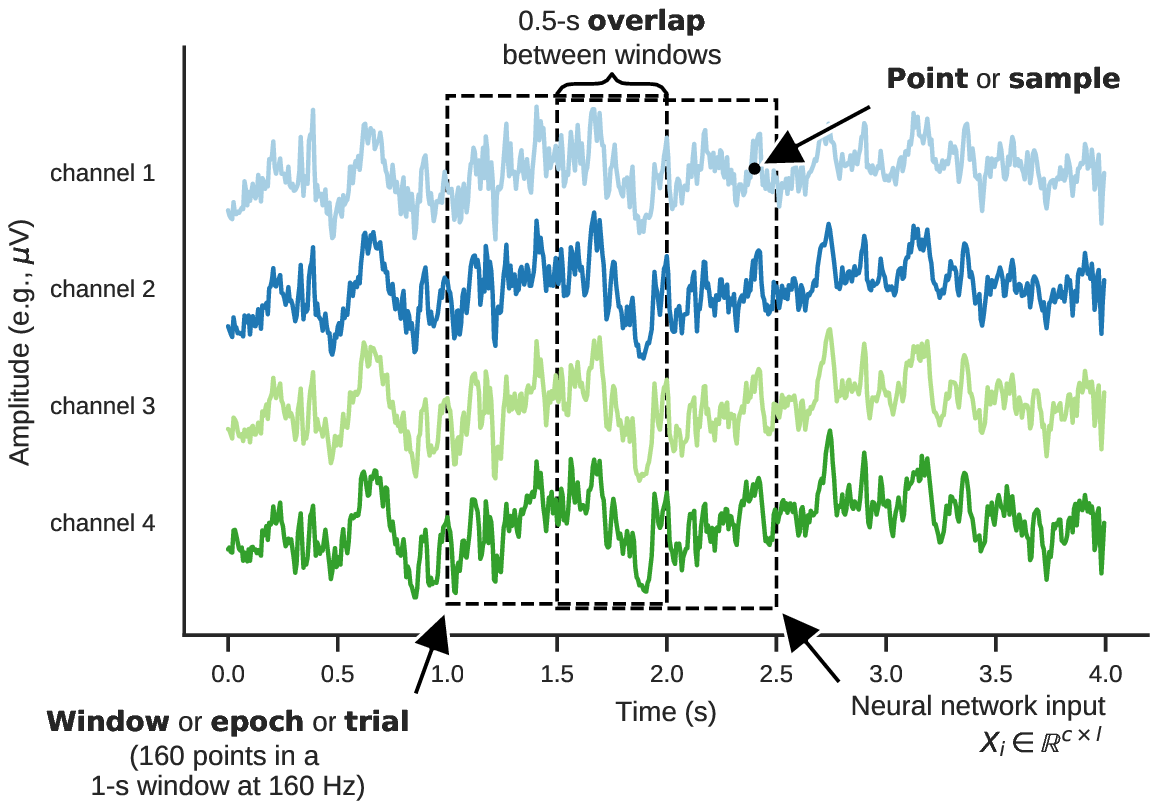}
    \caption{Overlapping windows (which may correspond to trials or epochs in some cases) are extracted from multichannel EEG recordings.}
    \label{fig:eeg}
    \end{subfigure}
    ~
    
    \begin{subfigure}{\textwidth}
    \begin{center}
    \begin{tikzpicture}[x=1.5cm, y=2cm, >=stealth, scale=0.8]
    \foreach \m/\l [count=\y] in {1,2,3,missing,4}
    \node [every neuron/.try, neuron \m/.try] (input-\m) at (0,2.5-\y) {};

    \foreach \m [count=\y] in {1,2,missing,3}
    \node [every neuron/.try, neuron \m/.try ] (hidden1-\m) at (1.5,2-\y*1.05) {};
  
    \foreach \m [count=\y] in {1,2,missing,3}
    \node [every neuron/.try, neuron \m/.try ] (hidden2-\m) at (4.5,2-\y*1.05) {};

    \foreach \m [count=\y] in {1,missing,2}
    \node [every neuron/.try, neuron \m/.try ] (output-\m) at (6,1.6-\y*1.1) {};

    \foreach \l [count=\i] in {1,2,3,n}
    \draw [<-] (input-\i) -- ++(-0.8,0)
    node [above, midway] {$x_\l$};
    
    \foreach \l [count=\i] in {1,2,n}
    \draw [->] (hidden1-\i) -- ++(0.8,0);
    
    \foreach \l [count=\i] in {1,2,n}
    \draw [<-] (hidden2-\i) -- ++(-0.8,0);

    \foreach \l [count=\i] in {1,m}
    \draw [->] (output-\i) -- ++(0.8,0)
    node [above, midway] {$y_\l$};

    \foreach \i in {1,...,4}
    \foreach \j in {1,...,3}
    \draw [->] (input-\i) -- (hidden1-\j);
    
    \foreach \i in {1,...,3}
    \foreach \j in {1,...,2}
    \draw [->] (hidden2-\i) -- (output-\j);
    
    \node [align=center, below] at (3,-0.45) {$\cdots$};

    \draw [decorate,decoration={brace,amplitude=0.3cm},rotate=90, xshift=-4.95cm,yshift=-8.2cm]
(0,0.7) -- (0,3) node [black,midway, below, yshift=-0.3cm] {\footnotesize Hidden layers};

    \draw [dashed,decorate,decoration={brace,amplitude=1.3cm}, xshift=-1.7cm,yshift=-6.3cm]
    (0,0.5) -- (0,4.8) node [black,midway,left, xshift=-1.5cm, align=left] 
    {\begin{varwidth}{3cm}{\footnotesize Example composed of points or samples (e.g. raw EEG, features)}\end{varwidth}};
    
    \draw [dashed,decorate,decoration={brace,amplitude=0.6cm},rotate=180, xshift=-10.5cm,yshift=-2.2cm] (0,0.5) -- (0,3) node [black,midway,right, xshift=0.8cm, align=left] {\begin{varwidth}{2.5cm}{\footnotesize Prediction (e.g. sleep stage, BCI classification)}\end{varwidth}};

    \foreach \l [count=\x from 0] in {Input \\ layer, , , Output \\ layer}
    \node [align=center, below] at (\x*2,-3) {\l};

    \end{tikzpicture}
    \end{center}
    \caption{Illustration of a general neural network architecture.}
    \label{fig:dl}
    \end{subfigure}
    \caption{Deep learning-based EEG processing pipeline and related terminology.}
    \label{fig:intro_dleeg}
\end{figure}

Some terms are sometimes used in the fields of machine learning, deep learning, statistics, EEG and signal processing with different meanings.
For example, in machine learning, ``sample'' usually refers to one example of the input received by a model, whereas in statistics, it can be used to refer to a group of examples taken from a population. It can also refer to the measure of a single time point in signal processing and EEG.
Similarly, in deep learning, the term ``epoch'' refers to one pass through the whole training set during training; in EEG, an epoch is instead a grouping of consecutive EEG time points extracted around a specific marker.
To avoid the confusion, we include in Table~\ref{tab:terminology} definitions for a few terms as used in this review. Fig.~\ref{fig:intro_dleeg} gives a visual example of what these terms refer to.

\begin{table}[]
\centering
\begin{tabular}{lp{0.5\textwidth}}
\toprule
                  & Definition used in this review                                                                                             \\
\hline
Point or sample   & A measure of the instantaneous electric potential picked up by the EEG sensors, typically in $\mu V$.                           \\
Example           & An instantiation of the data received by a model as input, typically denoted by $\mathbf{x}_i$ in the machine learning literature. \\
Trial             & A realization of the task under study, e.g., the presentation of one image in a visual ERP paradigm.                                           \\
Window or segment & A group of consecutive EEG samples extracted for further analysis, typically between 0.5 and 30 seconds.                   \\
Epoch             & A window extracted around a specific trial. \\
\bottomrule
\end{tabular}
\caption{Disambiguation of common terms used in this review.}
\label{tab:terminology}
\end{table}

\subsection{Objectives of the review}

This systematic review covers the current state-of-the-art in \gls{dl}-based \gls{eeg} processing by analyzing a large number of recent publications.
It provides an overview of the field for researchers familiar with traditional \gls{eeg} processing techniques and who are interested in applying \gls{dl} to their data.
At the same time, it aims to introduce the field applying \gls{dl} to \gls{eeg} to \gls{dl} researchers interested in expanding the types of data they benchmark their algorithms with, or who want to contribute to \gls{eeg} research.
For readers in any of these scenarios, this review also provides detailed methodological information on the various components of a DL-EEG pipeline to inform their own implementation\footnote{Additional information with more fine-grained data can be found in our data items table available at \url{http://dl-eeg.com}.}.
In addition to reporting trends and highlighting interesting approaches, we distill our analysis into a few recommendations in the hope of fostering reproducible and efficient research in the field.

\subsection{Organization of the review}

The review is organized as follows: Section~\ref{sec:introduction} briefly introduces key concepts in \gls{eeg} and \gls{dl}, and details the aims of the review; Section~\ref{sec:methods} describes how the systematic review was conducted, and how the studies were selected, assessed and analyzed; Section~\ref{sec:results} focuses on the most important characteristics of the studies selected and describes trends and promising approaches; Section~\ref{sec:discussion} discusses critical topics and challenges in DL-EEG, and provides recommendations for future studies; and Section~\ref{sec:conclusion} concludes by suggesting future avenues of research in DL-EEG. 
Finally, supplementary material containing our full data collection table, as well as the code used to produce the graphs, tables and results reported in this review, are made available online.



\section{Methods} \label{sec:methods}


English journal and conference papers, as well as electronic preprints, published between January 2010 and July 2018, were chosen as the target of this review. PubMed, Google Scholar and arXiv were queried to collect an initial list of papers to be reviewed. \footnote{The queries used for each database are available at \url{http://dl-eeg.com}.}
Additional papers were identified by scanning the reference sections of these papers.
The databases were queried for the last time on July 2, 2018.



The following title and abstract search terms were used to query the databases:
\begin{enumerate*}
\item \label{itm:search1} EEG,
\item \label{itm:search2} electroencephalogra*,
\item \label{itm:search3} deep learning,
\item \label{itm:search4} representation learning,
\item \label{itm:search5} neural network*,
\item \label{itm:search6} convolutional neural network*,
\item \label{itm:search7} ConvNet,
\item \label{itm:search8} CNN,
\item \label{itm:search9} recurrent neural network*,
\item \label{itm:search10} RNN,
\item \label{itm:search11} long short-term memory,
\item \label{itm:search12} LSTM,
\item \label{itm:search13} generative adversarial network*,
\item \label{itm:search14} GAN,
\item \label{itm:search15} autoencoder,
\item \label{itm:search16} restricted boltzmann machine*,
\item \label{itm:search17} deep belief network* and
\item \label{itm:search18} DBN
\end{enumerate*}.
The search terms were further combined with logical operators in the following way: 
(\ref{itm:search1} OR \ref{itm:search2}) AND (\ref{itm:search3} OR \ref{itm:search4} OR \ref{itm:search5} OR \ref{itm:search6} OR \ref{itm:search7} OR \ref{itm:search8} OR \ref{itm:search9} OR \ref{itm:search10} OR \ref{itm:search11} OR \ref{itm:search12} OR \ref{itm:search13} OR \ref{itm:search14} OR \ref{itm:search15} OR \ref{itm:search16} OR \ref{itm:search17} OR \ref{itm:search18}).
The papers were then included or excluded based on the criteria listed in Table~\ref{tab:inclusion_exclusion_criteria}.

\begin{table}
\centering
\caption{Inclusion and exclusion criteria.}
{\begin{tabular}{p{7cm}p{7cm}}
\toprule
Inclusion criteria & Exclusion criteria \\
\hline
\begin{itemize}[leftmargin=*]
\item Training of one or multiple deep learning architecture(s) to process non-invasive EEG data. 
\end{itemize} &
\begin{itemize}[leftmargin=*]
\item Studies focusing solely on invasive EEG (e.g., \gls{ecog} and intracortical EEG) or \gls{meg}.
\item Papers focusing solely on software tools.
\item Review articles.
\end{itemize} \\
\bottomrule
\end{tabular}}
\label{tab:inclusion_exclusion_criteria}
\end{table}

To assess the eligibility of the selected papers, the titles were read first. If the title did not clearly indicate whether the inclusion and exclusion criteria were met, the abstract was read as well. Finally, when reading the full text during the data collection process, papers that were found to be misaligned with the criteria were rejected.

Non-peer reviewed papers, such as arXiv electronic preprints\footnote{\url{https://arxiv.org/}}, are a valuable source of state-of-the-art information as their release cycle is typically shorter than that of peer-reviewed publications. Moreover, unconventional research ideas are more likely to be shared in such repositories, which improves the diversity of the reviewed work and reduces the bias possibly introduced by the peer-review process \cite{paez2017gray}. Therefore, non-peer reviewed preprints were also included in our review.
However, whenever a peer-reviewed publication followed a preprint submission, the peer-reviewed version was used instead.

A data extraction table was designed containing different data items relevant to our research questions, based on previous reviews with similar scopes and the authors' prior knowledge of the field. Following a first inspection of the papers with the data extraction sheet, data items were added, removed and refined. 
Each paper was initially reviewed by a single author, and then reviewed by a second if needed.
For each article selected, around 70 data items were extracted covering five categories: origin of the article, rationale, data used, EEG processing methodology, DL methodology and reported results. Table~\ref{table_data_items} lists and defines the different items included in each of these categories.
We make this data extraction table openly available for interested readers to reproduce our results and dive deeper into the data collected.
We also invite authors of published work in the field of DL and EEG to contribute to the table by verifying its content or by adding their articles to it.

The first category covers the origin of the article, that is whether it comes from a journal, a conference publication or a preprint repository, as well as the country of the first author's affiliation.
This gives a quick overview of the types of publication included in this review and of the main actors in the field.
Second, the rationale category focuses on the domains of application of the selected studies.
This is valuable information to understand the extent of the research in the field, and also enables us to identify trends across and within domains in our analysis.
Third, the data category includes all relevant information on the data used by the selected papers.
This comprises both the origin of the data and the data collection parameters, in addition to the amount of data that was available in each study. 
Through this section, we aim to clarify the data requirements for using DL on EEG.
Fourth, the EEG processing parameters category highlights the typical transformations required to apply DL to EEG, and covers preprocessing steps, artifact handling methodology, as well as feature extraction.
Fifth, details of the DL methodology, including architecture design, training procedures and inspection methods, are reported to guide the interested reader through state-of-the-art techniques.
Sixth, the reported results category reviews the results of the selected articles, as well as how they were reported, and aims to clarify how DL fares against traditional processing pipelines performance-wise. 
Finally, the reproducibility of the selected articles is quantified by looking at the availability of the data and code.
The results of this section support the critical component of our discussion.

\begin{table}
\centering
\caption{Data items extracted for each article selected.}
\scriptsize
{\begin{tabular}{p{0.1\textwidth}p{0.2\textwidth}p{0.6\textwidth}}
\toprule
Category & Data item & Description \\
\hline
Origin of article & 
Type of publication &
Whether the study was published as a journal article, a conference paper or in an electronic preprint repository.
\\
 & Venue &
Publishing venue, such as the name of a journal or conference.
\\
 & Country of first author affiliation & 
Location of the affiliated university, institute or research body of the first author.
\\
\hline
Study rationale & 
Domain of application &
Primary area of application of the selected study. In the case of multiple domains of application, the domain that was the focus of the study was retained.
\\
\hline
Data &
Quantity of data & 
Quantity of data used in the analysis, reported both in total number of samples and total minutes of recording.
\\
 & Hardware &
Vendor and model of the EEG recording device used.  
\\
 & Number of channels &
Number of EEG channels used in the analysis. May differ from the number of recorded channels.
\\
 & Sampling rate &
Sampling rate (reported in Hertz) used during the EEG acquisition.
\\
 & Subjects &
Number of subjects used in the analysis. May differ from the number of recorded subjects.
\\
 & Data split and cross-validation &
Percentage of data used for training, validation, and test, along with the cross-validation technique used, if any.
\\
 & Data augmentation &
Data augmentation technique used, if any, to generate new examples.
\\
\hline
EEG processing
 & Preprocessing &
Set of manipulation steps applied to the raw data to prepare it for use by the architecture or for feature extraction.
\\
 & Artifact handling &
Whether a method for cleaning artifacts was applied.
\\
 & Features &
Output of the feature extraction procedure, which aims to better represent the information of interest contained in the preprocessed data.
\\
\hline
\multirow{2}{1.4cm}{Deep learning methodology} 
 & Architecture &
Structure of the neural network in terms of types of layers (e.g. fully-connected, convolutional).
\\
 & Number of layers &
Measure of architecture depth.
\\
 & EEG-specific design choices &
Particular architecture choices made with the aim of processing EEG data specifically. 
\\
 & Training procedure &
Method applied to train the neural network (e.g., standard optimization, unsupervised pre-training followed by supervised fine-tuning, etc.).
\\
 & Regularization &
Constraint on the hypothesis class intended to improve a learning algorithm generalization performance  (e.g., weight decay, dropout). 
\\
 & Optimization &
Parameter update rule.
\\
 & Hyperparameter search &
Whether a specific method was employed in order to tune the hyperparameter set.  
\\
 & Subject handling &
Intra- vs inter-subject analysis.
\\
 & Inspection of trained models &
Method used to inspect a trained DL model.
\\
\hline
Results &
Type of baseline &
Whether the study included baseline models that used traditional processing pipelines, DL baseline models, or a combination of the two. 
\\
 & Performance metrics &
Metrics used by the study to report performance (e.g., accuracy, f1-score, etc.).
\\
 & Validation procedure &
Methodology used to validate the performance of the trained models, including cross-validation and data split.
\\
 & Statistical testing &
Types of statistical tests used to assess the performance of the trained models.
\\
 & Comparison of results &
Reported results of the study, both for the trained DL models and for the baseline models.
\\
\hline
Reproducibility &
Dataset &
Whether the data used for the experiment comes from private recordings or from a publicly available dataset.
\\
 & Code &
Whether the code used for the experiment is available online or not, and if so, where.
\\
\bottomrule
\end{tabular}}
\label{table_data_items}
\end{table}




\section{Results} \label{sec:results}

\begin{figure}[h]
\includegraphics[width=0.5\textwidth]{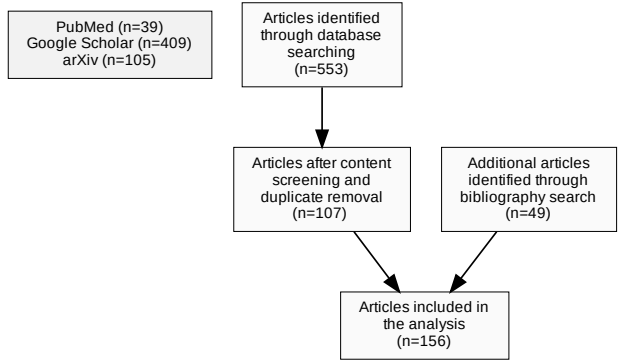}
\centering
\caption{Selection process for the papers.}
\label{fig_prisma_diagram}
\end{figure}

The database queries yielded 553 different results that matched the search terms (see  Fig.~\ref{fig_prisma_diagram}). 49 additional papers were then identified using the reference sections of the initial papers. Based on our inclusion and exclusion criteria, 446 papers were excluded. Therefore, 156 papers were selected for inclusion in the analysis.

\subsection{Origin of the selected studies}

\begin{figure}[h]
\includegraphics[width=1\textwidth]{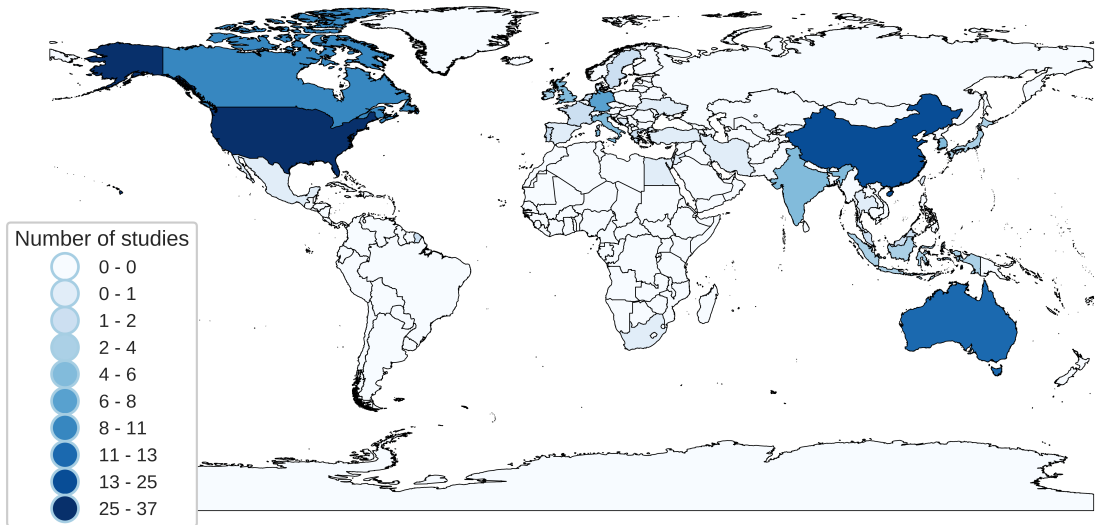}
\caption{Countries of first author affiliations.}
\label{fig:countrymap}
\end{figure}

Our search methodology returned $49$ journal papers, $58$ conference and workshop papers, $48$ preprints and $1$ journal paper supplement (\cite{Yan2018}, included in the "Journal" category in our analysis) that met our criteria.
A total of $23$ journal and conference papers had initially been made available as preprints on arXiv or bioRxiv.
Popular journals included \textit{Neurocomputing}, \textit{Journal of Neural Engineering} and \textit{Biomedical Signal Processing and Control}, each with three publications contained in our selected studies.
We also looked at the location of the first author's affiliation to get a sense of the geographical distribution of research on DL-EEG. We found that most contributions came from the USA, China and Australia (see Fig.~\ref{fig:countrymap}).


\subsection{Domains} \label{sec:domains}

\begin{figure}
\includegraphics[width=\linewidth]{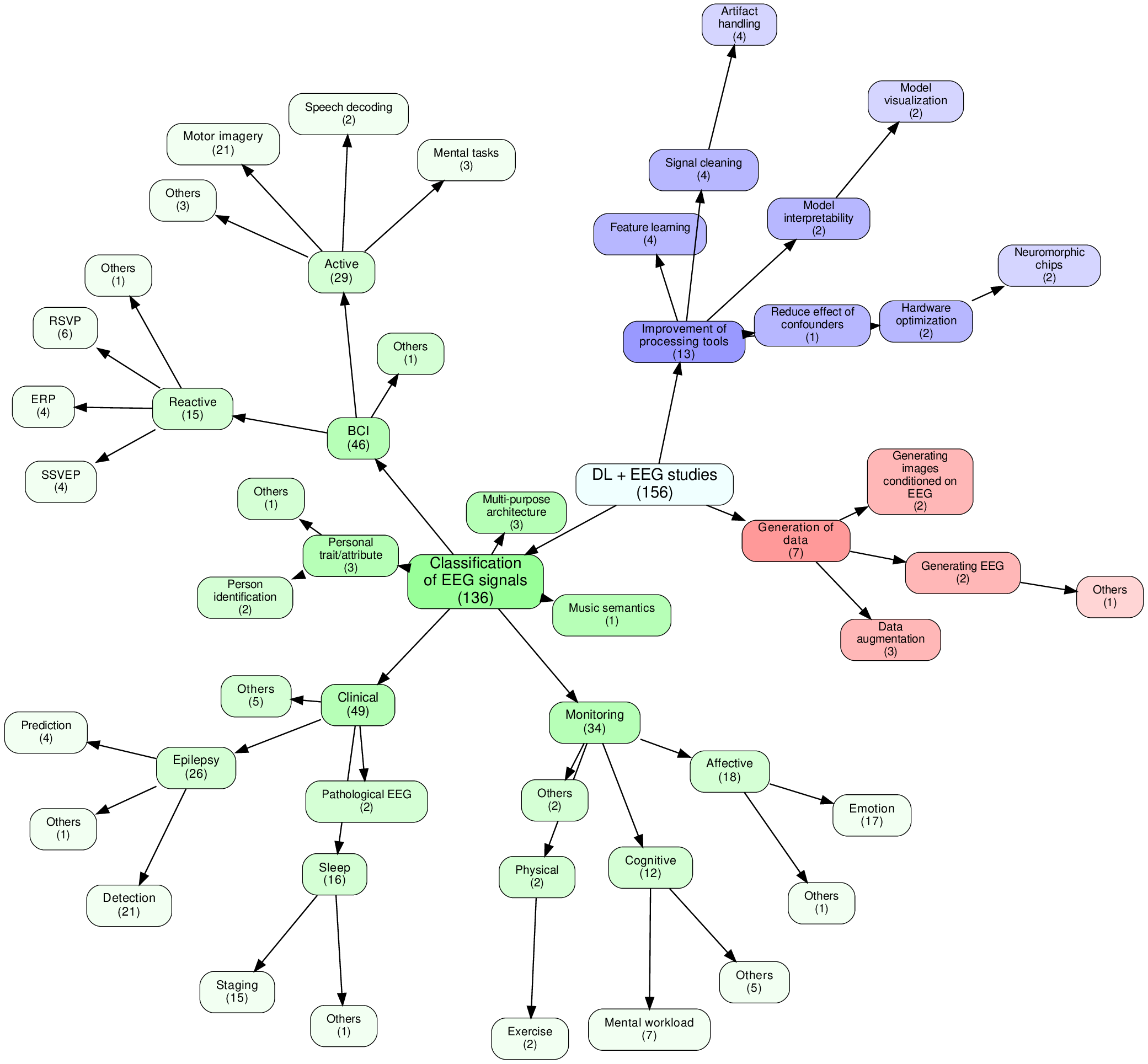}
\centering
\caption{Focus of the studies. The number of papers that fit in a category is showed in brackets for each category. Studies that covered more than one topic were categorized based on their main focus.}
\label{fig:domaintree}
\end{figure}

The selected studies applied DL to EEG in various ways (see Fig.~\ref{fig:domaintree} and Table~\ref{tab:domains_vs_architectures}). Most studies ($86\%$) focused on using DL for the classification of EEG data, most notably for sleep staging, seizure detection and prediction, brain-computer interfaces (BCIs), as well as for cognitive and affective monitoring. Around $9\%$ of the studies focused instead on the improvement of processing tools, such as learning features from EEG, handling artifacts, or visualizing trained models. The remaining papers ($5\%$) explored ways of generating data from EEG, e.g. augmenting data, or generating images conditioned on EEG.

\begin{figure}[t]
\includegraphics[width=0.5\textwidth]{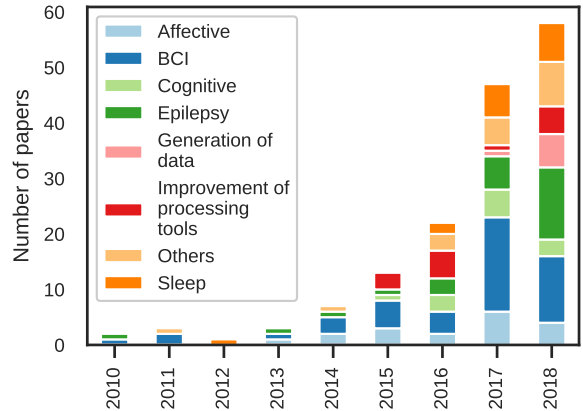}
\centering
\caption{Number of publications per domain per year. To simplify the figure, some of the categories defined in Fig.~\ref{fig:domaintree} have been grouped together.}
\label{fig:domainyears}
\end{figure}

Despite the absolute number of DL-EEG publications being relatively small as compared to other DL applications such as computer vision~\cite{lecun2015deep}, there is clearly a growing interest in the field. Fig.~\ref{fig:domainyears} shows the growth of the DL-EEG literature since 2010.
The first seven months of 2018 alone count more publications than 2010 to 2016 combined, hence the relevance of this review. It is, however, still too early to conclude on trends concerning the application domains, given the relatively small number of publications to date.

\begin{sidewaystable}
    \centering
    \caption{Categorization of the selected studies according to their application domain and DL architecture. Domains are divided into four levels, as described in Fig.~\ref{fig:domaintree}.}
    \resizebox{\textwidth}{!}{
    \begin{tabular}{p{0.15\textheight}p{0.15\textheight}p{0.15\textheight}p{0.15\textheight}c>{\centering}m{0.15\textheight}ccccccccc}
    \hline
Domain 1 & Domain 2 & Domain 3 & Domain 4 &                                                          AE &                                                                                                                         CNN &                                                                CNN+RNN &                                  DBN &                                                         FC &                  GAN &                     MLP &                           N/M &               Other &                     RBM &                                                                  RNN \\
\midrule
Classification of EEG signals & BCI &   & Detection &                                                             &                                                                                                                             &                                                                        &                                      &                                                            &                      &  \cite{drouin2016using} &                               &                     &                         &                                                                      \\
                                &                 & Active & Grasp and lift &                                                             &                                                                                                                             &                                                                        &                                      &                                                            &                      &                         &                               &                     &                         &                                                        \cite{An2016} \\
                                &                 &                   & Mental tasks &                                                             &                                                                                                                             &                                                                        &                                      &                                       \cite{Padmanabh2017} &                      &                         &                               &                     &                         &                                        \cite{Hasib2018, Patnaik2017} \\
                               &                 &                   & Motor imagery &                                   \cite{Zhang2017a, Li2014} &  \cite{Gao2018, Sakhavi2017, Tang2017, Schirrmeister2017, Dharamsi2017, Loshchilov2017, Tabar2016a, Sakhavi2015, Yang2015a} &                                                      \cite{Zhang2018c} &                        \cite{An2014} &  \cite{Chiarelli2018, Major2017, Alomari2013, Mohamed2011} &                      &                         &            \cite{Normand2015} &   \cite{Zhang2017g} &                         &                                            \cite{Zhang2017d, Bu2010} \\
                              &                 &                   & RSVP &                                                             &                                                                                                         \cite{Shamwell2016} &                                                                        &                                      &                                                            &                      &                         &                               &                     &                         &                                                                      \\
                                &                 &                   & Slow cortical potentials &                                                             &                                                                                                                             &                                                                        &                                      &                                                            &                      &                         &               \cite{Ding2015} &                     &                         &                                                                      \\
                                &                 &                   & Speech decoding &                                                             &                                                                                                                             &                                                                        &                      \cite{Sree2017} &                                                            &                      &                         &                               &                     &          \cite{Sun2016} &                                                                      \\
                                &                 & Active \& Reactive & MI \& ERP &                                                             &                                                                                                          \cite{Lawhern2018} &                                                                        &                                      &                                                            &                      &                         &                               &                     &                         &                                                                      \\
                                &                 & Reactive & ERP &                                                             &                                                                       \cite{Yoon2018, Volker2018, Behncke2017, Cecotti2011} &                                                                        &                                      &                                                            &                      &                         &                               &                     &                         &                                                                      \\
                                &                 &                   & Heard speech decoding &                                                             &                                                                                                                             &                                                                        &                                      &                                                            &                      &                         &                               &                     &                         &                                                \cite{Moinnereau2018} \\
                                &                 &                   & RSVP &                                                             &                                                       \cite{Parekh2018, Zafar2017, Hajinoroozi2017, Manor2015, Cecotti2014} &                                                  \cite{Spampinato2017} &                                      &                                                            &                      &                         &                               &                     &                         &                                                                      \\
                                &                 &                   & SSVEP &                                    \cite{Perez-Benitez2018} &                                                                                   \cite{Aznan2018, Waytowich2018, kwak2017} &                                                                        &                                      &                                                            &                      &                         &                               &                     &                         &                                                                      \\
                                & Clinical & Alzheimer's disease &   &                                                             &                                                                                                         \cite{Morabito2016} &                                                                        &                                      &                                                            &                      &                         &                               &                     &                         &                                                                      \\
                                &                 & Anomaly detection &   &                                                             &                                                                                                                             &                                                                        &                    \cite{Wulsin2011} &                                                            &                      &                         &                               &                     &                         &                                                                      \\
                                &                 & Dementia &   &                                         \cite{Morabito2017} &                                                                                                                             &                                                                        &                                      &                                                            &                      &                         &                               &                     &                         &                                                                      \\
                                &                 & Epilepsy & Detection &                          \cite{Yuan2018a, Golmohammadi2017} &                                         \cite{Yan2018, Hao2018, Ullah2018, Shea2018, Oshea2017, Acharya2017, Thodoroff2016} &  \cite{VanPutten2018a, Golmohammadi2017b, Shah2017, Golmohammadi2017a} &                    \cite{Turner2014} &                          \cite{Pramod2015, Omerhodzic2013} &                      &                         &               \cite{Taqi2017} &                     &                         &  \cite{Hussein2018, Ahmedt-Aristizabal2018, Talathi2017, Naderi2010} \\
                                &                 &                   & Event annotation &                                                             &                                                                                                                             &                                                                        &                                      &                                                            &                      &                         &              \cite{Yang2016b} &                     &                         &                                                                      \\
                                &                 &                   & Prediction &                                                             &                                                                                                 \cite{Truong2018, Page2016} &                                                                        &                                      &                                                            &                      &                         &                               &  \cite{Truong2018a} &                         &                                                  \cite{Tsiouris2018} \\
                                &                 & Ischemic stroke &   &                                                             &                                                                                                            \cite{Giri2016a} &                                                                        &                                      &                                                            &                      &                         &                               &                     &                         &                                                                      \\
                                &                 & Pathological EEG &   &                                                             &                                                                                                   \cite{Schirrmeister2017a} &                                                         \cite{Roy2018} &                                      &                                                            &                      &                         &                               &                     &                         &                                                                      \\
                                &                 & Schizophrenia & Detection &                                                             &                                                                                                              \cite{Chu2017} &                                                                        &                                      &                                                            &                      &                         &                               &                     &                         &                                                                      \\
                                &                 & Sleep & Abnormality detection &                                                             &                                                                                                         \cite{Ruffini2018a} &                                                                        &                                      &                                                            &                      &                         &                               &                     &                         &                                                                      \\
                                &                 &                   & Staging &                          \cite{Langkvist2018, Tripathy2018} &                    \cite{Patanaik2018, Phan2018, Sors2018, Chambon2018, Vilamala2017, Xie2017, Manzano2017a, Tsinalis2016a} &                                        \cite{Supratak2017, Biswal2017} &                 \cite{Langkvist2012} &                                                            &                      &                         &                               &                     &                         &                                       \cite{dong2018mixed, Giri2016} \\
                                & Monitoring & Affective & Bullying incidents &                                                             &                                                                                                        \cite{Baltatzis2017} &                                                                        &                                      &                                                            &                      &                         &                               &                     &                         &                                                                      \\
                                &                 &                   & Emotion &  \cite{BenSaid2017a, Xu2016, Liu2016, Jirayucharoensak2014} &                                                                                                    \cite{Liao2018, Lin2017} &                                                                        &  \cite{Zheng2015, Zheng2014, Li2013} &                             \cite{Teo2018, Frydenlund2015} &                      &                         &  \cite{Mehmood2017, Kwon2017} &                     &      \cite{Gao2015} &                                \cite{Li2018, Zhang2018, Alhagry2017} \\
                                &                 & Cognitive & Drowsiness &                                                             &                                                                                                                             &                                                                        &                                      &                                                            &                      &                         &                               &                     &  \cite{Hajinoroozi2015} &                                                                      \\
                                &                 &                   & Engagement &                                                             &                                                                                                                             &                                                                        &                                      &                                                            &                      &                         &                               &                     &           \cite{Li2017} &                                                                      \\
                                &                 &                   & Eyes closed/open &                                                             &                                                                                                                             &                                                                        &                    \cite{Narejo2016} &                                                            &                      &                         &                               &                     &                         &                                                                      \\
                                &                 &                   & Fatigue &                                                             &                                                                                                      \cite{Hajinoroozi2016} &                                                                        &                                      &                                                            &                      &                         &                               &                     &                         &                                                                      \\
                                &                 &                   & Mental workload &                                   \cite{Yin2017a, Yin2016b} &                                                                                              \cite{Almogbel2018, Zhang2017} &                                          \cite{Hefron2018, Kuanar2018} &                                      &                                                            &                      &                         &                               &                     &                         &                                                    \cite{Hefron2017} \\
                                &                 &                   & Mental workload \& fatigue &                                                             &                                                                                                                             &                                                                        &                       \cite{Yin2017} &                                                            &                      &                         &                               &                     &                         &                                                                      \\
                                &                 & Cognitive vs. Affective &   &                                                             &                                                                                                                             &                                                                        &                 \cite{Bashivan2016b} &                                                            &                      &                         &                               &                     &                         &                                                                      \\
                                &                 & Music semantics &   &                                                             &                                                                                                           \cite{Stober2014} &                                                                        &                                      &                                                            &                      &                         &                               &                     &                         &                                                                      \\
                                &                 & Physical & Exercise &                                                             &                                                                                                            \cite{Ghosh2018} &                                                                        &                                      &                                                            &                      &                         &          \cite{Gordienko2017} &                     &                         &                                                                      \\
                                & Multi-purpose architecture &   &   &                                             \cite{Lee2018a} &                                                                                                           \cite{Zhang2018a} &                                                                        &                                      &                                                            &                      &                         &                               &    \cite{Deiss2018} &                         &                                                                      \\
                                & Music semantics &   &   &                                                             &                                                                                                           \cite{Raposo2017} &                                                                        &                                      &                                                            &                      &                         &                               &                     &                         &                                                                      \\
                                & Personal trait/attribute & Person identification &   &                                                             &                                                                                                                             &                                                                        &                                      &                                                            &                      &                         &                               &                     &                         &                                        \cite{Zhang2017e, Zhang2017c} \\
                                &                 & Sex &   &                                                             &                                                                                                       \cite{VanPutten2018b} &                                                                        &                                      &                                                            &                      &                         &                               &                     &                         &                                                                      \\
\midrule
Generation of data & Data augmentation &   &   &                                                             &                                                                                 \cite{Wang2018, Zhang2018b, Schwabedal2018} &                                                                        &                                      &                                                            &                      &                         &                               &                     &                         &                                                                      \\
                                & Generating EEG &   &   &                                                             &                                                                                                                             &                                                                        &                                      &                                                            &  \cite{Hartmann2018} &                         &                               &                     &                         &                                                                      \\
                                &                 & Spatial upsampling &   &                                                             &                                                                                                                             &                                                                        &                                      &                                                            &    \cite{Corley2018} &                         &                               &                     &                         &                                                                      \\
                                & Generating images conditioned on EEG &   &   &                                                             &                                                                                                                             &                                                                        &                                      &                                                            &       \cite{Lee2018} &                         &                               &  \cite{Palazzo2017} &                         &                                                                      \\
\midrule
Improvement of processing tools & Feature learning &   &   &                          \cite{Wen2018, Stober2015, Li2015} &                                                                                                        \cite{Bashivan2016a} &                                                                        &                                      &                                                            &                      &                         &                               &                     &                         &                                                                      \\
                                & Hardware optimization & Neuromorphic chips &   &                                                             &                                                                                                            \cite{Nurse2016} &                                                                        &                                      &                                                            &                      &                         &              \cite{Yepes2017} &                     &                         &                                                                      \\
                                & Model interpretability & Model visualization &   &                                                             &                                                                                                        \cite{Hartmann2018b} &                                                                        &                                      &                                                            &                      &                         &              \cite{Sturm2016} &                     &                         &                                                                      \\
                                & Reduce effect of confounders &   &   &                                                             &                                                                                                                             &                                                                        &                                      &                                                            &                      &                         &                               &                     &                         &                                                        \cite{Wu2018} \\
                                & Signal cleaning & Artifact handling &   &                                  \cite{Yang2018, Yang2016a} &                                                                                                            \cite{Wang2018a} &                                                                        &                                      &                                                            &                      &                         &                               &                     &                         &                                                   \cite{Pardede2015} \\ \hline
    \end{tabular}}
    \label{tab:domains_vs_architectures}
\end{sidewaystable}


\subsection{Data} \label{sec:data}

The availability of large datasets containing unprecedented numbers of examples is often mentioned as one of the main enablers of deep learning research in the early 2010s \cite{goodfellow2016deep}. It is thus crucial to understand what the equivalent is in EEG research, given the relatively high cost of collecting EEG data. Given the high dimensionality of EEG signals \cite{lotte2018review}, one would assume that a considerable amount of data is required. Although our analysis cannot answer that question fully, we seek to cover as many dimensions of the answer as possible to give the reader a complete view of what has been done so far.

\subsubsection{Quantity of data}

We make use of two different measures to report the amount of data used in the reviewed studies: 1) the number of examples available to the deep learning network and 2) the total duration of the EEG recordings used in the study, in minutes. Both measures include the EEG data used across training, validation and test phases. For an in-depth analysis of the amount of data, please see the data items table which contains more detailed information.

\begin{figure}[h]
    \includegraphics[width=\textwidth]{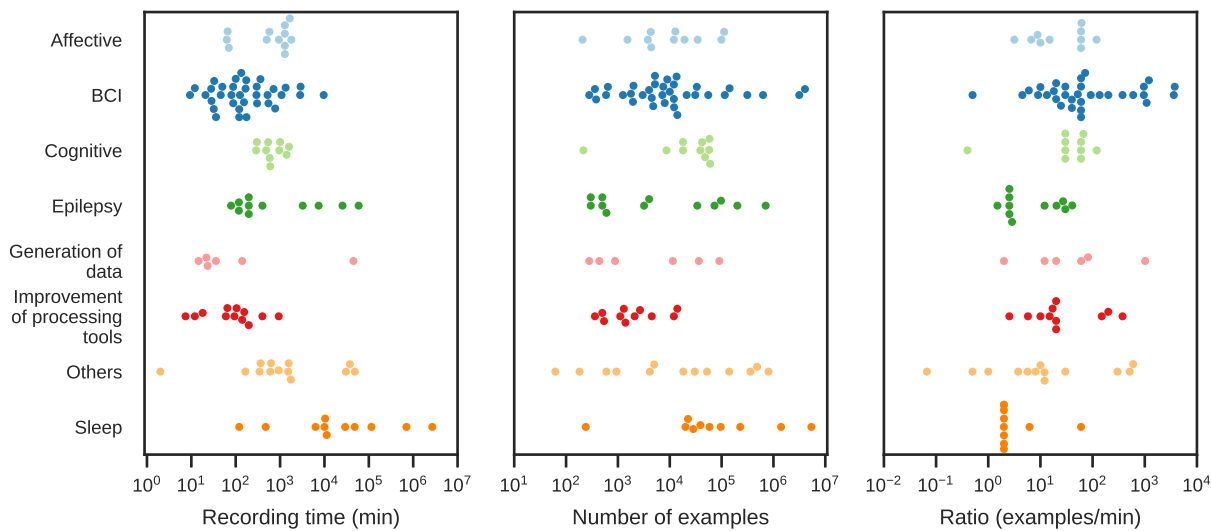}
    \centering
    \caption{Amount of data used by the selected studies. Each dot represents one dataset. The left column shows the datasets according to the total length of the EEG recordings used, in minutes. The center column shows the number of examples that were extracted from the available EEG recordings. The right column presents the ratio of number of examples to minutes of EEG recording.}
    \label{fig:data}
\end{figure}

The left column of Fig.~\ref{fig:data} shows the amount of EEG data, in minutes, used in the analysis of each study, including training, validation and/or testing. Therefore, the time reported here does not necessarily correspond to the total recording time of the experiment(s). For example, many studies recorded a baseline at the beginning and/or at the end but did not use it in their analysis. Moreover, some studies recorded more classes than they used in their analysis. Also, some studies used sub-windows of recorded epochs (e.g. in a motor imagery BCI, using 3~s of a 7~s epoch). The amount of data in minutes used across the studies ranges from 2 up to 4,800,000 (mean = 62,602; median = 360).

The center column of Fig.~\ref{fig:data} shows the amount of examples available to the models, either for training, validation or test. This number presents a relevant variability as some studies used a sliding window with a significant overlap generating many examples (e.g., 250~ms windows with 234~ms overlap, therefore generating 4,050,000 examples from 1080~minutes of EEG data \cite{Shamwell2016}), while some other studies used very long windows generating very few examples (e.g., 15-min windows with no overlap, therefore generating 62 examples from 930~minutes of EEG data \cite{Giri2016a}). The wide range of windowing approaches (see Section~\ref{ssec:data_augmentation}) indicates that a better understanding of its impact is still required. The number of examples used ranged from 62 up to 9,750,000 (mean = 251,532; median = 14,000).

The right column of Fig.~\ref{fig:data} shows the ratio between the amount of data in minutes and the number of examples. This ratio was never mentioned specifically in the papers reviewed but we nonetheless wanted to see if there were any trends or standards across domains and we found that in sleep studies for example, this ratio tends to be of two as most people are using 30~s non-overlapping windows. Brain-computer interfacing is seeing the most sparsity perhaps indicating a lack of best practices for sliding windows. It is important to note that the BCI field is also the one in which the exact relevant time measures were hardest to obtain since most of the recorded data isn't used (e.g. baseline, in-between epochs). Therefore, some of the sparsity on the graph could come from us trying our best to understand and calculate the amount of data used (i.e., seen by the model). Obviously, in the following categories: generation of data, improvement of processing tools and others, this ratio has little to no value as the trends would be difficult to interpret.

The amount of data across different domains varies significantly. In domains like sleep and epilepsy, EEG recordings last many hours (e.g., a full night), but in domains like affective and cognitive monitoring, the data usually comes from lab experiments on the scale of a few hours or even a few minutes.

\subsubsection{Subjects} \label{ssec:subjects}

\begin{figure}[h]
    \includegraphics[width=0.666\textwidth]{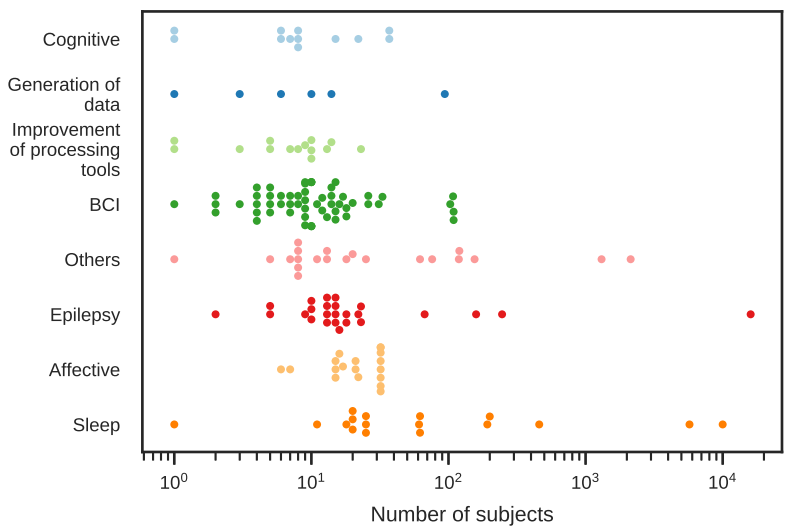}
    \centering
    \caption{Number of subjects per domain in datasets. Each point represents one dataset used by one of the selected studies.}
    \label{fig:data_nb_subjects}
\end{figure}

Often correlated with the amount of data, the number of subjects also varies significantly across studies (see Fig.~\ref{fig:data_nb_subjects}). 
Half of the datasets used in the selected studies contained fewer than 13 subjects. Six studies, in particular, used datasets with a much greater number of subjects: \cite{Patanaik2018, Sors2018, VanPutten2018b, Schirrmeister2017a} all used datasets with at least 250 subjects, while \cite{Biswal2017} and \cite{Golmohammadi2017} used datasets with 10,000 and 16,000 subjects, respectively. As explained in Section~\ref{ss:intra_inter_subject}, the untapped potential of DL-EEG might reside in combining data coming from many different subjects and/or datasets to train a model that captures common underlying features and generalizes better. In \cite{Yang2018}, for example, the authors trained their model using an existing public dataset and also recorded their own EEG data to test the generalization on new subjects. In \cite{Volker2018}, an increase in performance was observed when using more subjects during training before testing on new subjects. The authors tested using from 1 to 30 subjects with a leave-one-subject-out cross-validation scheme, and reported an increase in performance with noticeable diminishing returns above 15 subjects.

\subsubsection{Recording parameters} 

\begin{figure}[h]
    \begin{subfigure}{0.5\textwidth}
    \includegraphics[width=0.9\textwidth]{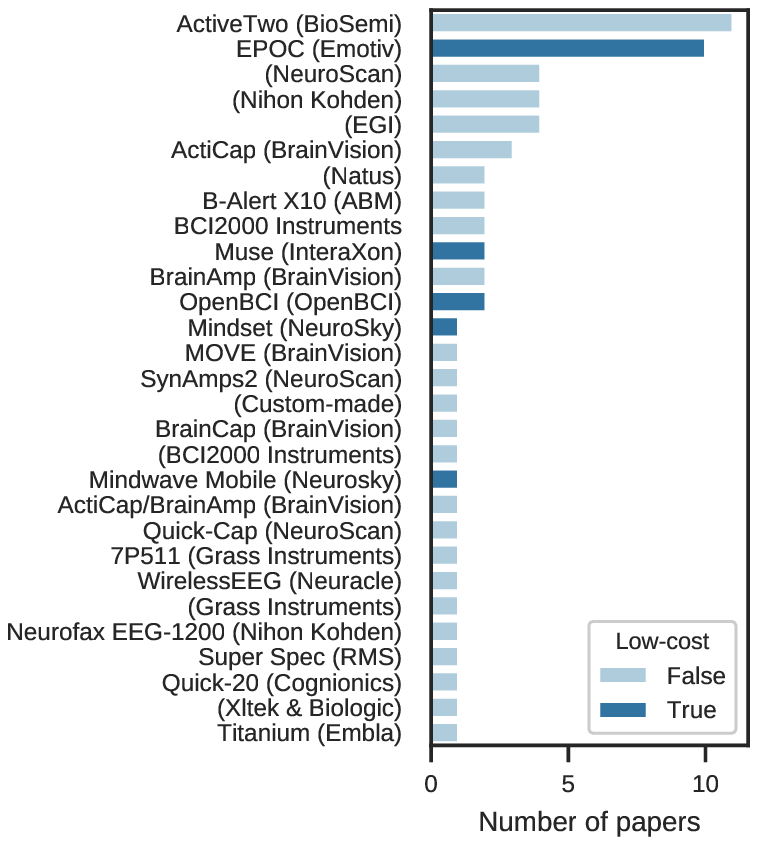}
    \caption{EEG hardware used in the studies. The device name is followed by the manufacturer's name in parentheses. Low-cost devices (defined as devices below \$1,000 excluding software, licenses and accessories) are indicated by a different color.}
    \label{fig:data_eeg_hardware}
    \end{subfigure}
    ~
    \begin{subfigure}{0.5\textwidth}
    \includegraphics[width=0.9\textwidth]{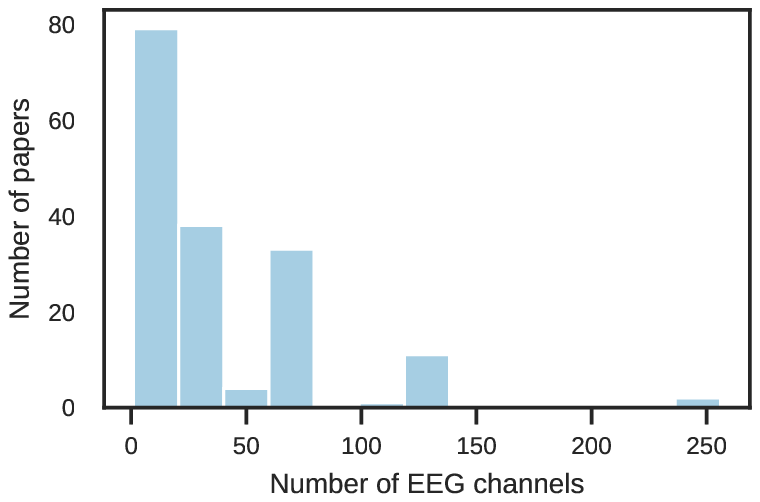}
    \caption{Distribution of the number of EEG channels.}
    \label{fig:data_nb_channels}
    \end{subfigure}
    
    \caption{Hardware characteristics of the EEG devices used to collect the data in the selected studies.}
    \label{fig:hardware}
\end{figure}

As shown later in Section~\ref{sec:reproducibility}, $42\%$ of reported results came from private recordings. We look at the type of EEG device that was used by the selected studies to collect their data, and additionally highlight low-cost, often called "consumer" EEG devices, as compared to traditional "research" or "medical" EEG devices (see Fig.~\ref{fig:data_eeg_hardware}). We loosely defined low-cost EEG devices as devices under the USD 1,000 threshold (excluding software, licenses and accessories). Among these devices, the Emotiv EPOC was used the most, followed by the OpenBCI, Muse and Neurosky devices. As for the research grade EEG devices, the BioSemi ActiveTwo was used the most, followed by BrainVision products.

The EEG data used in the selected studies was recorded with 1 to 256 electrodes, with half of the studies using between 8 and 62 electrodes (see Fig.~\ref{fig:data_nb_channels}). The number of electrodes required for a specific task or analysis is usually arbitrarily defined as no fundamental rules have been established. In most cases, adding electrodes will improve possible analyses by increasing spatial resolution. However, adding an electrode close to other electrodes might not provide significantly different information, while increasing the preparation time and the participant's discomfort and requiring a more costly device. Higher density EEG devices are popular in research but hardly ecological. In \cite{Shah2017}, the authors explored the impact of the number of channels on the specificity and sensitivity for seizure detection. They showed that increasing the number of channels from 4 up to 22 (including two referential channels) resulted in an increase in sensitivity from $31\%$ to $39\%$ and from $40\%$ to $90\%$ in specificity. They concluded, however, that the position of the referential channels is very important as well, making it difficult to compare across datasets coming from different neurologists and recording sites using different locations for the reference(s) channel(s).

Similarly, in \cite{Chambon2018}, the impact of different electrode configurations was assessed on a sleep staging task.
The authors found that increasing the number of electrodes from two to six produced the highest increase in performance, while adding additional sensors, up to 22 in total, also improved the performance but not as much.
The placement of the electrodes in a 2-channel montage also impacted the performance, with central and frontal montages leading to better performance than posterior ones on the sleep staging task.

Furthermore, EEG sampling rates varied mostly between 100 and 1000~Hz in the selected studies (the sampling rate reported here is the one used to record the EEG data and not after downsampling, as described in Section~\ref{sec:preprocessing}). Around $50\%$ of studies used sampling rates of 250~Hz or less and the highest sampling rate used was 5000~Hz (\cite{Hartmann2018b}).

\subsubsection{Data augmentation} \label{ssec:data_augmentation}

Data augmentation is a technique by which new data examples are artificially generated from the existing training data. Data augmentation has proven efficient in other fields such as computer vision, where data manipulations including rotations, translations, cropping and flipping can be applied to generate more training examples~\cite{perez2017}. Adding more training examples allows the use of more complex models comprising more parameters while reducing overfitting. When done properly, data augmentation increases accuracy and stability, offering a better generalization on new data \cite{zhang2016understanding}.

Out of the 156 papers reviewed, three papers explicitly explored the impact of data augmentation on DL-EEG (\cite{Wang2018, Zhang2018b, Schwabedal2018}). Interestingly, each one looked at it from the perspective of a different domain: sleep, affective monitoring and BCI. Also, all three are from 2018, perhaps showing an emerging interest in data augmentation.
First, in \cite{Wang2018}, Gaussian noise was added to the training data to obtain new examples.
This approach was tested on two different public datasets for emotion classification (SEED \cite{zheng2018emotionmeter} and MAHNOB-HCI \cite{soleymani2012multimodal}). 
They improved their accuracy on the SEED dataset using LeNet (\cite{lecun1998gradient}) from $49.6\%$ (without augmentation) to $74.3\%$ (with augmentation), from $34.2\%$ (without) to $75.0\%$ (with) using ResNet (\cite{he2016deep}) and from $40.8\%$ (without) to $45.4\%$ (with) on MAHNOB-HCI dataset using ResNet. 
Their best accuracy was obtained with a standard deviation of 0.2 and by augmenting the data to 30 times its original size. Despite impressive results, it is important to note that they also compared LeNet and ResNet to an SVM which had an accuracy of $74.2\%$ (without) and $73.4\%$ (with) on the SEED dataset. This might indicate that the initial amount of data was insufficient for LeNet or ResNet but adding data clearly helped bring the performance up to par with the SVM. 
Second, in \cite{Zhang2018b}, a conditional deep convolutional generative adversarial network (cDCGAN) was used to generate artificial EEG signals on one of the BCI Competition motor imagery datasets. 
Using a CNN, it was shown that data augmentation helped improve accuracy from $83\%$ to around $86\%$ to classify motor imagery.
In \cite{Schwabedal2018}, the authors explicitly targeted the class imbalance problem of under-represented sleep stages by generating Fourier transform (FT) surrogates of raw EEG data on the CAPSLPDB dataset. They improved their accuracy up to $24\%$ on some classes.

An additional 30 papers explicitly used data augmentation in one form or another but only a handful investigated the impact it hae on performance.
In \cite{Kuanar2018, Bashivan2016a}, noise was added to 2D feature images, although it did not improve results in \cite{Bashivan2016a}.
In \cite{Hussein2018}, artifacts such as eye blinks and muscle activity, as well as Gaussian white noise, were used to augment the data and improve robustness.
In \cite{Yin2017} and \cite{Yin2017a}, Gaussian noise was added to the input feature vector. This approach increased the accuracy of the SDAE model from around $76.5\%$ (without augmentation) to $85.5\%$ (with). 

Multiple studies also used overlapping windows as a way to augment their data, although many did not explicitly frame this as data augmentation.
In \cite{Ullah2018, Oshea2017}, overlapping windows were explicitly used as a data augmentation technique. In \cite{kwak2017}, different shift lengths between overlapping windows (from 10 ms to 60 ms out of a 2-s window) were compared, showing that by generating more training samples with smaller shifts, performance improved significantly. 
In \cite{Schirrmeister2017}, the concept of overlapping windows was pushed even further: 1) redundant computations due to EEG samples being in more than one window were simplified thanks to "cropped training", which ensured these computations were only done once, thereby speeding up training and 2) the fact that overlapping windows share information was used to design an additional term to the cost function, which further regularizes the models by penalizing decisions that are not the same while being close in time.

Other procedures used the inherent spatial and temporal characteristics of EEG to augment their data.
In \cite{Deiss2018}, the authors doubled their data by swapping the right and left side electrodes, claiming that as the task was a symmetrical problem, which side of the brain expresses the response would not affect classification. 
In \cite{BenSaid2017a}, the authors augmented their multimodal (EEG and EMG) data by duplicating samples and keeping the values from one modality only, while setting the other modality values to 0 and vice-versa. 
In \cite{Frydenlund2015}, the authors made use of the data that is usually thrown away when downsampling EEG in the preprocessing stage. It is common to downsample a signal acquired at higher sampling rate to 256~Hz or less. In their case, they reused the data thrown away during that step as new samples: a downsampling by a factor of $N$ would therefore allow an augmentation of $N$ times.

Finally, classification of rare events where the number of available samples are orders of magnitude smaller than their counterpart classes \cite{Schwabedal2018} is another motivation for data augmentation.
In EEG classification, epileptic seizures or transitional sleep stages (e.g. S1 and S3) often lead to such unbalanced classes.
In \cite{Vilamala2017}, the class imbalance problem was addressed by randomly balancing all classes while sampling for each training epoch.
Similarly, in \cite{Chambon2018}, balanced accuracy was maximized by using a balanced sampling strategy.
In \cite{Tsiouris2018}, EEG segments from the interictal class were split into smaller subgroups of equal size to the preictal class. 
In \cite{Sors2018}, cost-sensitive learning and oversampling were used to solve the class imbalance problem for sleep staging but the overall performance using these approaches did not improve.
In \cite{Ruffini2018a}, the authors randomly replicated subjects from the minority class to balance classes. 
Similarly, in \cite{Supratak2017, dong2018mixed, drouin2016using, Manor2015}, oversampling of the minority class was used to balance classes.
Conversely, in \cite{Thodoroff2016, Shamwell2016}, the majority class was subsampled. 
In \cite{Truong2018}, an overlapping window with a subject-specific overlap was used to match classes. Similar work by the same group \cite{Truong2018a} showed that when training a GAN on individual subjects, augmenting data with an overlapping window increased accuracy from $60.91\%$ to $74.33\%$.
For more on imbalanced learning, we refer the interested reader to \cite{haixiang2017learning}.


\subsection{EEG processing} \label{sec:preprocessing}

One of the oft-claimed motivation for using deep learning on \gls{eeg} processing is automatic feature learning \cite{Patanaik2018, Hussein2018, Ghosh2018, Hasib2018, Moinnereau2018, Aznan2018, Zafar2017}. This can be explained by the fact that feature engineering is a time-consuming task \cite{li2015eeg}. Additionally, preprocessing and cleaning \gls{eeg} signals from artifacts is a demanding step of the usual EEG processing pipeline. Hence, in this section, we look at aspects related to data preparation, such as preprocessing, artifact handling and feature extraction. This analysis is critical to clarify what level of preprocessing \gls{eeg} data requires to be successfully used with deep neural networks. 

\subsubsection{Preprocessing}

Preprocessing \gls{eeg} data usually comprises a few general steps, such as downsampling, band-pass filtering, and windowing. Throughout the process of reviewing papers, we found that a different number of preprocessing steps were employed in the studies. In \cite{Hefron2018}, it is mentioned that ``a substantial amount of preprocessing was required'' for assessing cognitive workload using DL. More specifically, it was necessary to trim the \gls{eeg} trials, downsample the data to 512~Hz and 64 electrodes, identify and interpolate bad channels, calculate the average reference, remove line noise, and high-pass filter the data starting at 1~Hz. On the other hand, Stober et al. \cite{Stober2015} applied a single preprocessing step by removing the bad channels for each subject. In studies focusing on emotion recognition using the DEAP dataset \cite{Koelstra2012}, the same preprocessing methodology proposed by the researchers that collected the dataset was typically used, i.e., re-referencing to the common average, downsampling to 256~Hz, and high-pass filtering at 2~Hz.

\begin{figure}[h]
\includegraphics[width=\textwidth]{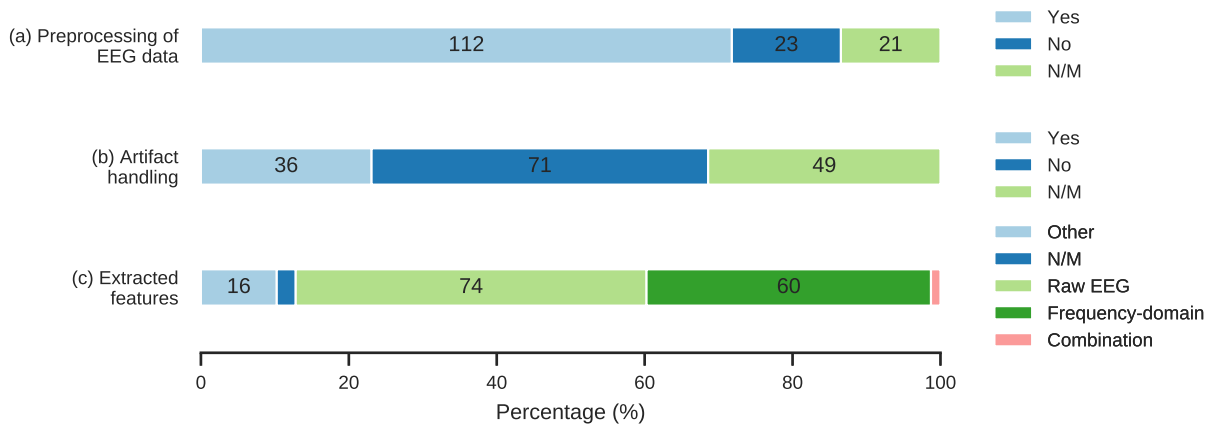}
\caption{\gls{eeg} processing choices. (a) Number of studies that used preprocessing steps, such as filtering, (b) number of studies that included, rejected or corrected artifacts in their data and (c) types of features that were used as input to the proposed models.}
\label{fig:preproc}
\end{figure}

We separated the papers into three categories based on whether or not they used preprocessing steps: ``Yes'', in cases where preprocessing was employed; ``No'', when the authors explicitly mentioned that no preprocessing was necessary; and not mentioned (``N/M'') when no information was provided. The results are shown in Fig.~\ref{fig:preproc}.

A considerable proportion of the reviewed articles ($72\%$) employed at least one preprocessing method such as downsampling or re-referencing. This result is not surprising, as applications of \glspl{dnn} to other domains, such as computer vision, usually require some kind of preprocessing like cropping and normalization as well.

\subsubsection{Artifact handling}

artifact handling techniques are used to remove specific types of noise, such as ocular and muscular artifacts \cite{uriguen2015eeg}. As emphasized in \cite{Yang2016a}, removal of artifacts may be crucial for achieving good \gls{eeg} decoding performance. Adding this to the fact that cleaning \gls{eeg} signals might be a time-consuming process, some studies attempted to apply only minimal preprocessing such as removing bad channels and leave the burden of learning from a potentially noisy signal on the neural network \cite{Stober2015}. With that in mind, we decided to look at artifact handling separately.

artifact removal techniques usually require the intervention of a human expert \cite{nolan2010faster}. Different techniques leverage human knowledge to different extents, and might fully rely on an expert, as in the case of visual inspection, or require prior knowledge to simply tune a hyperparameter, as in the case of wavelet-based \gls{ica}  \cite{makeig1996independent}. Among the studies which handled artifacts, a myriad of techniques were applied. Some studies employed methods which rely on human knowledge such as amplitude thresholding \cite{Moinnereau2018}, manual identification of high-variance segments \cite{Hefron2018}, and handling \gls{eeg} blinking-related noise based on high-amplitude EOG segments \cite{Manor2015}. On the other hand, many other articles favored techniques that rely less on human intervention, such as blind source separation techniques. For instance, in \cite{Sun2016, Yin2016b, Yin2017a, Ghosh2018, Parekh2018, Patnaik2017}, \gls{ica} was used to separate ocular components from \gls{eeg} data.

In order to investigate the necessity of removing artifacts from \gls{eeg} when using deep neural networks, we split the selected papers into three categories, in a similar way to the preprocessing analysis (see Fig.~\ref{fig:preproc}).  
Almost half the papers ($46\%$) did not use artifact handling methods, while $24\%$ did. Additionally, $31\%$ of the studies did not mention whether artifact handling was necessary to achieve their results. Given those results, we are encouraged to believe that using \glspl{dnn} on \gls{eeg} might be a way to avoid the explicit artifact removal step of the classical \gls{eeg} processing pipeline without harming task performance.

\subsubsection{Features}


Feature engineering is one of the most demanding steps of the traditional \gls{eeg} processing pipeline \cite{li2015eeg} and the main goal of many papers considered in this review \cite{Patanaik2018, Hussein2018, Ghosh2018, Hasib2018, Moinnereau2018, Aznan2018, Zafar2017} is to get rid of this step by employing deep neural networks for automatic feature learning. This aspect appears to be of interest to researchers in the field since its early stages, as indicated by the work of Wulsin et al. \cite{Wulsin2011}, which, in 2011, compared the performance of \glspl{dbn} on classification and anomaly detection tasks using both raw \gls{eeg} and features as inputs. More recently, studies such as \cite{Sturm2016, Hartmann2018} achieved promising results without the need to extract features. 

On the other hand, a considerable proportion of the reviewed papers used hand-engineered features as the input to their deep neural networks. In \cite{Teo2018}, for example, authors used a time-frequency domain representation of \gls{eeg} obtained via the \gls{stft} for detecting binary user-preference (like versus dislike). Similarly, Truong et al. \cite{Truong2018}, used the \gls{stft} as a 2-dimensional \gls{eeg} representation for seizure prediction using \glspl{cnn}. In \cite{Zhang2017}, \gls{eeg} frequency-domain information was also used. Widely adopted by the \gls{eeg} community, the \gls{psd} of classical frequency bands from around 1~Hz to 40~Hz were used as features. Specifically, authors selected the delta (1-4~Hz), theta (5-8~Hz), alpha (9-13~Hz), lower beta (14-16~Hz), higher beta (17-30~Hz), and gamma (31-40~Hz) bands for mental workload state recognition. Moreover, other studies employed a combination of features, for instance \cite{Giri2016a}, which used PSD features, as well as entropy, kurtosis, fractal component, among others, as input of the proposed \gls{cnn} for ischemic stroke detection.

Given that the majority of \gls{eeg} features are obtained in the frequency-domain, our analysis consisted in separating the reviewed articles into four categories according to the respective input type. Namely, the categories were: ``Raw \gls{eeg}'', ``Frequency-domain'', ``Combination'' (in case more than one type of feature was used), and ``Other'' (for papers using neither raw \gls{eeg} nor frequency-domain features). Studies that did not specify the type of input were assigned to the category ``N/M'' (not mentioned). Notice that, here, we use ``feature'' and ``input type'' interchangeably.   

Fig.~\ref{fig:preproc} presents the result of our analysis. One can observe that $49\%$ of the papers used only raw \gls{eeg} data as input, whereas $48\%$ used hand-engineered features, from which $36\%$ corresponded to frequency domain-derived features. Finally, $3\%$ did not specify the type of input of their model. According to these results, we find indications that \glspl{dnn} can be in fact applied to raw \gls{eeg} data and achieve state-of-the-art results.


\subsection{Deep learning methodology} \label{sec:dl_methodology}

\subsubsection{Architecture} \label{ssec:architecture}

A crucial choice in the DL-based EEG processing pipeline is the neural network architecture to be used. In this section, we aim at answering a few questions on this topic, namely: 1) "What are the most frequently used architectures?", 2) "How has this changed across years?", 3) "Is the choice of architecture related to input characteristics?" and 4) "How deep are the networks used in DL-EEG?". 

\begin{figure}[h]
    \begin{subfigure}[b]{0.33\textwidth}
    \includegraphics[width=\linewidth]{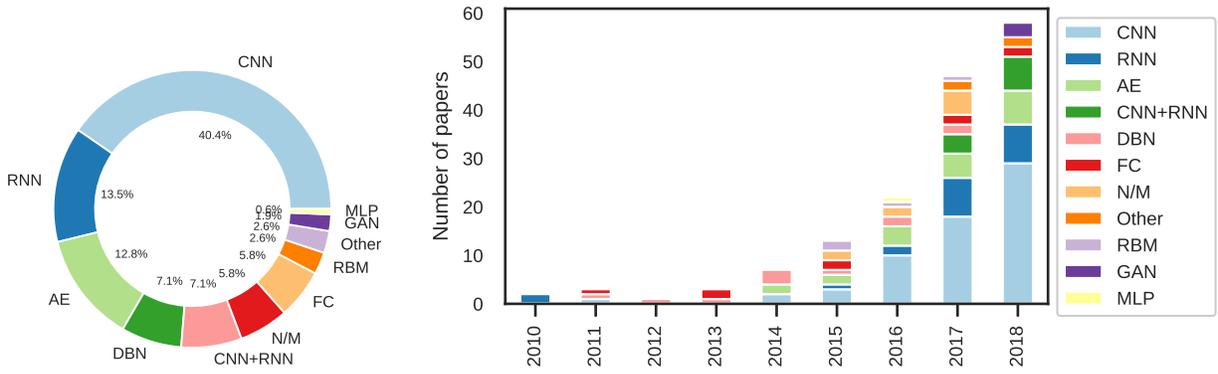}
    \caption{Architectures.}
    \label{fig:arch_total}
    \end{subfigure}
    ~
    \begin{subfigure}[b]{0.66\textwidth}
    \includegraphics[width=\linewidth]{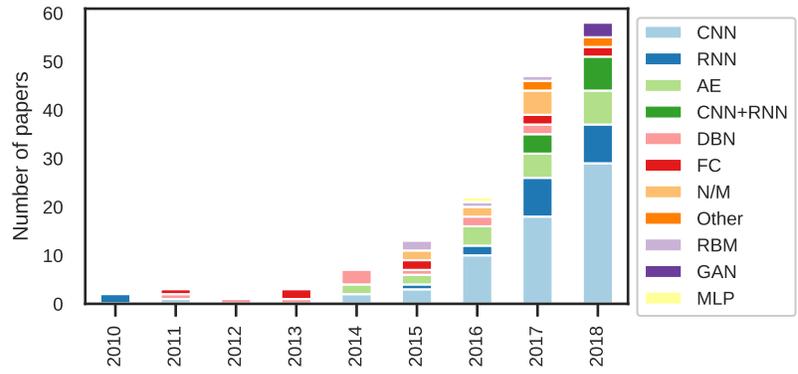}
    \caption{Distribution of architectures across years.}
    \label{fig:arch_years}
    \end{subfigure}
    
    \begin{subfigure}[b]{0.5\textwidth}
    \centering
    \includegraphics[width=\linewidth]{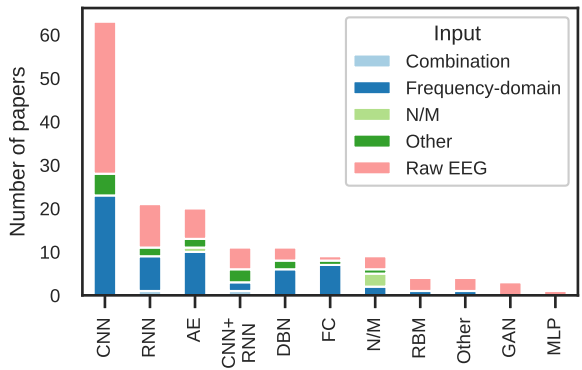}
    \caption{Distribution of input type according to the architecture category.}
    \label{fig:arch_input}
    \end{subfigure}
    ~
    \begin{subfigure}[b]{0.5\textwidth}
    \centering
    \includegraphics[width=\linewidth]{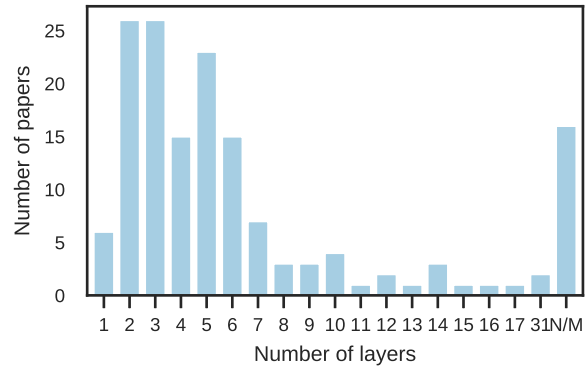}
    \caption{Distribution of number of neural network layers.}
    \label{fig:number_layers}
    \end{subfigure}
    
    \caption{Deep learning architectures used in the selected studies. ``N/M'' stands for ``Not mentioned'' and accounts for papers which have not reported the respective deep learning methodology aspect under analysis.}
    \label{fig:arch}
\end{figure}

To answer the first three questions, we divided and assigned the architectures used in the 156 papers into the following groups: \glspl{cnn}, \glspl{rnn}, \glspl{ae}, \glspl{rbm}, \glspl{dbn}, \glspl{gan}, \gls{fc} networks, combinations of \glspl{cnn} and \glspl{rnn} (CNN+RNN), and ``Others'' for any other architecture or combination not included in the aforementioned categories.
Fig.~\ref{fig:arch_total} shows the percentage of studies that used the different architectures. $41\%$ of the papers used \glspl{cnn}, whereas \glspl{rnn} and \glspl{ae} were the architecture choice of about $14\%$ and $13\%$ of the works, respectively. Combinations of CNNs and RNNs, on the other hand, were used in $7\%$ of the studies. RBMs and DBNs corresponded together to $10\%$ of the architectures. FC neural networks were employed by $6\%$ of the papers. GANs and other architectures appeared in $5\%$ of the considered cases. Notice that $6\%$ of the analyzed papers did not report their choice of architecture.

In Fig.~\ref{fig:arch_years}, we provide a visualization of the distribution of architecture types across years. Until the end of 2014, DBNs and FC networks comprised the majority of the studies. However, since 2015, CNNs have been the architecture type of choice in most studies. This can be attributed to the their capabilities of end-to-end learning and of exploiting hierarchical structure on the data \cite{schirrmeister2017deep}, as well as their success and subsequent popularity on computer vision tasks, such as the ILSVRC 2012 challenge \cite{deng2012ilsvrc}. Interestingly, we also observe that as the number of papers grows, the proportion of studies using CNNs and combinations of recurrent and convolutional layers has been growing steadily. The latter shows that RNNs are increasingly of interest for EEG analysis.
On the other hand, the use of architectures such as RBMs, DBNs and AEs has been decreasing with time. Commonly, models employing these architectures utilize a two-step training procedure consisting of 1) unsupervised feature learning and 2) training a classifier on top of the learned features. However, we notice that recent studies leverage the hierarchical feature learning capabilities of CNNs to achieve end-to-end supervised feature learning, i.e., training both a feature extractor and a classifier simultaneously. 

To complement the previous result, we cross-checked the architecture and input type information provided in Fig.~\ref{fig:preproc}. Results are presented in Fig.~\ref{fig:arch_input} and clearly show that CNNs are indeed used more often with raw EEG data as input. This corroborates the idea that researchers employ this architecture with the aim of leveraging the capabilities of deep neural networks to process EEG data in an end-to-end fashion, avoiding the time-consuming task of extracting features. From this figure, one can also notice that some architectures such as deep belief networks are typically used with frequency-domain features as inputs, while GANs, on the other hand, have been only applied to EEG processing using raw data.

\paragraph{Number of layers}

Deep neural networks are usually composed of stacks of layers which provide hierarchical processing. Although one might think the use of \textit{deep} neural networks implies the existence of a large number of layers in the architecture, there is no absolute consensus in the literature regarding this definition. Here we investigate this aspect and show that the number of layers is not necessarily large, i.e., larger than three, in many of the considered studies. 
 
In Fig~\ref{fig:number_layers}, we show the distribution of the reviewed papers according to the number of layers in the respective architecture. For studies reporting results for different architectures and number of layers, we only considered the highest value. We observed that most of the selected studies (128) utilized architectures with at most 10 layers. A total of 16 articles have not reported the architecture depth. When comparing the distribution of papers according to the architecture depth with architectures commonly used for computer vision applications, such as VGG-16 (16 layers) \cite{simonyan2014very} and ResNet-18 (18 layers) \cite{he2016deep}, we observe that the current literature on DL-EEG suggests shallower models achieve better performance. 


Some studies specifically investigated the effect of increasing the model depth. Zhang et al. \cite{Zhang2017} evaluated the performance of models with depth ranging from two to 10 on a mental workload classification task. Architectures with seven layers outperformed both shallower (two and four layers) and deeper (10 layers) models in terms of accuracy, precision, F-measure and G-mean. Moreover, O'Shea et al. \cite{Oshea2017} compared the performance of a \gls{cnn} with six and 11 layers on neonatal seizure detection. Their results show that, in this case, the deeper network presented better \gls{rocauc} in comparison to the shallower model, as well as a \gls{svm}. In \cite{kwak2017}, the effect of depth on \gls{cnn} performance was also studied. The authors compared results obtained by a CNN with two and three convolutional layers on the task of classifying SSVEPs under ambulatory conditions. The shallower architecture outperformed the three-layer one in all scenarios considering different amounts of training data. \Gls{cca} together with a KNN classifier were also evaluated and employed as a baseline method. Interestingly, as the number of training samples increased, the shallower model outperformed the \gls{cca}-based baseline.

\paragraph{EEG-specific design choices}

Particular choices regarding the architecture might enable a model to mimic the process of extracting EEG features. An architecture can also be specifically designed to impose specific properties on the learned representations. This is for instance the case with max-pooling, which is used to produce invariant feature maps to slight translations on the input \cite{goodfellow2016deep}. In the case of EEG signals, one might be interested in forcing the model to process temporal and spatial information separately in the earlier stages of the network. In \cite{Chambon2018, kwak2017, Zafar2017, Behncke2017, Schirrmeister2017, Manor2015}, one-dimensional convolutions were used in the input layer with the aim of processing either temporal or spatial information independently at this point of the hierarchy. Other studies \cite{Zhang2017g, Supratak2017} combined recurrent and convolutional neural networks as an alternative to the previous approach of separating temporal and spatial content. Recurrent models were also applied in cases where it was necessary to capture long-term dependencies from the \gls{eeg} data \cite{Li2018, Zhang2018}.

\subsubsection{Training} \label{ssec:training}

Details regarding the training of the models proposed in the literature are of great importance as different approaches and hyperparameter choices can greatly impact the performance of neural networks. The use of pre-trained models, regularization, and hyperparameter search strategies are examples of aspects we took into account during the review process. We report our main findings in this section.

\paragraph{Training Procedure}
One of the advantages of applying deep neural networks to \gls{eeg} processing is the possibility of simultaneously training a feature extractor and a model for executing a downstream task such as classification or regression. However, in some of the reviewed studies \cite{Langkvist2018, Wen2018, Morabito2017}, these two tasks were executed separately. Usually, the feature learning was done in an unsupervised fashion, with \glspl{rbm}, \glspl{dbn}, or \glspl{ae}. After training those models to provide an appropriate representation of the EEG input signal, the new features were then used as the input for a target task which is, in general, classification. In other cases, pre-trained models were used for a different purpose, such as object recognition, and were fine-tuned on the specific \gls{eeg} task with the aim of providing a better initialization or regularization effect \cite{li2018explicit}.

In order to investigate the training procedure of the reviewed papers, we classify each one according to the adopted training procedure. Models which have parameters learned without using any kind of pre-training were assigned to the ``Standard'' group. The remaining studies, which specified the training procedure, were included in the ``Pre-training'' class, in case the parameters were learned in more than one step. Finally, papers employing different methodologies for training, such as co-learning \cite{Deiss2018}, were included in the ``Other'' group. 

\begin{figure}[h]
\centering
\includegraphics[width=\textwidth]{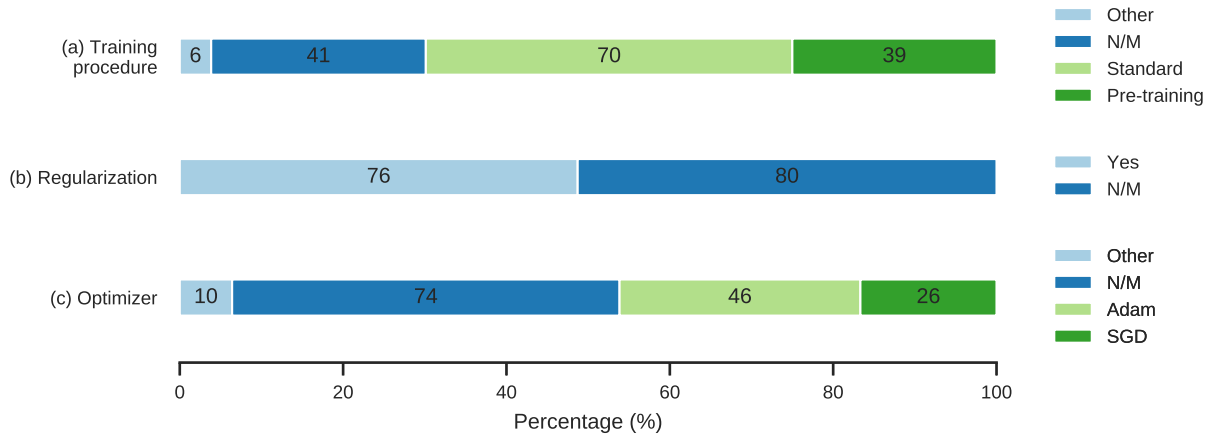}
\caption{Deep learning methodology choices. (a) Training methodology used in the studies, (b) number of studies that reported the use of regularization methods such as dropout, weight decay, etc. and (c) type of optimizer used in the studies.}
\label{fig:training_aspects}
\end{figure}

In Fig.~\ref{fig:training_aspects}a) we show how the reviewed papers are distributed according to the training procedure. ``N/M'' refers to studies which have not reported this aspect. Almost half the papers did not employ any pre-traning strategy, while $25\%$ did. Even though the training strategy is crucial for achieving good performance with deep neural networks, $26\%$ of the selected studies have not explicitly described it in their paper. 

\paragraph{Regularization}
In the context of our literature review, we define regularization as any constraint on the set of possible functions parametrized by the neural network intended to improve its performance on unseen data during training \cite{goodfellow2016deep}. The main goal when regularizing a neural network is to control its complexity in order to obtain better generalization performance \cite{bishop1995neural}, which can be verified by a decrease on test error in the case of classification problems. There are several ways of regularizing neural networks, and among the most common are weight decay (L2 and L1 regularization) \cite{goodfellow2016deep}, early stopping \cite{prechelt1998automatic}, dropout \cite{srivastava2014dropout}, and label smoothing \cite{szegedy2016rethinking}. Notice that even though the use of pre-trained models as initialization can also be interpreted as a regularizer \cite{li2018explicit}, in this work we decided to include it in the training procedure analysis instead.

As the use of regularization might be fundamental to guarantee a good performance on unseen data during training, we analyzed how many of the reviewed studies explicitly stated that they have employed it in their models. Papers were separated in two groups, namely: ``Yes'' in case any kind of regularization was used, and ``N/M'' otherwise. In Fig.~\ref{fig:training_aspects} we present the proportion of studies in each group. 

From Fig.~\ref{fig:training_aspects}, one can notice that more than half the studies employed at least one regularization method. Furthermore, regularization methods were frequently combined in the reviewed studies. Hefron et al. \cite{Hefron2018} employed a combination of dropout, L1- and L2- regularization to learn temporal and frequency representations across different participants. The developed model
was trained for recognizing mental workload states elicited by the MATB task \cite{comstock1992multi}. Similarly, Längkvist and Loutfi \cite{Langkvist2018}, combined two types of regularization with the aim of developing a model tailored to an automatic sleep stage classification task. Besides L2-regularization, they added a penalty term to encourage weight sparsity, defined as the KL-divergence between the mean activation of each hidden unit over all training examples in a training batch and a hyperparameter $\rho$.

\paragraph{Optimization}
Learning the parameters of a deep neural network is, in practice, an optimization problem. The best way to tackle it is still an open research question in the deep learning literature, as there is often a compromise between finding a good solution in terms of minimizing the cost function and the performance of a local optimum expressed by the generalization gap, i.e. the difference between the training error and the true error estimated on the test set. In this scenario, the choice of a parameter update rule, i.e. the \textit{learning algorithm} or \textit{optimizer}, might be key for achieving good results.

The most commonly used optimizers are reported in Fig.~\ref{fig:training_aspects}. One surprising finding is that even though the choice of optimizer is a fundamental aspect of the \gls{dl}-\gls{eeg} pipeline, $47\%$ of the considered studies did not report which parameter update rule was applied. Moreover, $30\%$ used Adam \cite{kingma2014adam} and $17\%$ Stochastic Gradient Descent \cite{robbins1985stochastic} (notice that we also refer to the mini-batch case as SGD). $6\%$ of the papers utilized different optimizers, such as RMSprop \cite{tieleman2012lecture}, Adagrad \cite{duchi2011adaptive}, and Adadelta \cite{zeiler2012adadelta}. 

Another interesting finding the optimizer analysis provided is the steady increase in the use of Adam. 
Indeed, from 2017 to 2018, the percentage of studies using Adam increased from $31.9\%$ to $52.6\%$. Adam was proposed as a gradient-based method with the capability of adaptively tuning the learning rate based on estimates of first and second order moments of the gradient. It became very popular in general deep neural networks applications (accumulating approximately 15,000 citations since 2014\footnote{Google scholar query run on 30/11/2018.}). 
Interestingly, we notice a proportional decrease from 2017 to 2018 of the number of papers which did not report the optimizer utilized. 


\paragraph{Hyperparameter search}
From a practical point-of-view, tuning the hyperparameters of a learning algorithm often takes up a great part of the time spent during training. \glspl{gan}, for instance, are known to be sensitive to the choices of optimizer and architecture hyperparameters \cite{gulrajani2017improved, li2017dualing}. In order to minimize the amount of time spent finding an appropriate set of hyperparameters, several methods have been proposed in the literature. Examples of commonly applied methods are grid search \cite{bergstra2012random} and Bayesian optimization \cite{snoek2012practical}. Grid search consists in determining a range of values for each parameter to be tuned, choosing values in this range, and evaluating the model, usually in a validation set considering all combinations. One of the advantages of grid search is that it is highly parallelizable, as each set of hyperparameter is independent of the other. Bayesian optimization, in turn, defines a posterior distribution over the hyperparameters space and iteratively updates its values according to the performance obtained by the model with a hyperparameter set corresponding to the expected posterior.

Given the importance of finding a good set of hyperparameters and the difficulty of achieving this in general, we calculate the percentage of papers that employed some search method for tuning their models and optimizers, as well as the amount of articles that have not included any information regarding this aspect. Results indicate that almost $80\%$ of the reviewed papers have not mentioned the use of hyperparameters search strategies. It is important to highlight that among those articles, it is not clear how many have not done any tuning at all and how many have just not considered to include this information in the paper. From the $21\%$ that declared to have searched for an appropriate set of hyperparameters, some have manually done this by trial and error (e.g. \cite{Acharya2017, dong2018mixed, Tsiouris2018, Patanaik2018}), while others employed grid search (e.g. \cite{Yin2016b, Xu2016, drouin2016using, Yin2017a, Liao2018, Aznan2018, Langkvist2018}), and a few used other strategies such as Bayesian methods (e.g. \cite{Stober2014, Stober2015, Schwabedal2018}). 


\subsection{Inspection of trained models} \label{ssec:visualization}

In this section, we review if, and how, studies have inspected their proposed models.
Out of the selected studies, $27\%$ reported inspecting their models.
Two studies focused more specifically on the question of model inspection in the context of DL and EEG \cite{Hartmann2018b, Ghosh2018}.
See Table~\ref{tab:model_inspection} for a list of the different techniques that were used by more than one study.
For a general review of DL model inspection techniques, see \cite{hohman2018visual}.

The most frequent model inspection techniques involved the analysis of the trained model's weights \cite{Perez-Benitez2018, Yoon2018, Langkvist2018, Deiss2018, Lawhern2018, Xu2016, Tsinalis2016a, Nurse2016, Tabar2016a, Zheng2015, Stober2015, Manor2015, Yang2015a}.
This often requires focusing on the weights of the first layer only, as their interpretation in regard to the input data is straightforward.
Indeed, the absolute value of a weight represents the strength with which the corresponding input dimension is used by the model - a higher value can therefore be interpreted as a rough measure of feature importance.
For deeper layers, however, the hierarchical nature of neural networks means it is much harder to understand what a weight is applied to.

The analysis of model activations was used in multiple studies \cite{Yuan2018a, Waytowich2018, Lawhern2018, kwak2017, Yin2017a, Supratak2017, Shamwell2016, Manor2015}.
This kind of inspection method usually involves visualizing the activations of the trained model over multiple examples, and thus inferring how different parts of the network react to known inputs.
The input-perturbation network-prediction correlation map technique, introduced in \cite{Schirrmeister2017a}, pushes this idea further by trying to identify causal relationships between the inputs and the decisions of a model.
The impact of the perturbation on the activations of the last layer's units then shines light onto which characteristics of the input are important for the classifier to make a correct prediction.
To do this, the input is first perturbed, either in the time- or frequency-domain, to alter its amplitude or phase characteristics \cite{Hartmann2018b}, and then fed into the network.
Occlusion sensitivity techniques \cite{Lee2018, Chambon2018, Thodoroff2016} use a similar idea, by which the decisions of the network when different parts of the input are occluded are analyzed.

Several studies used backpropagation-based techniques to generate input maps that maximize activations of specific units \cite{VanPutten2018b, Ruffini2018a, Sors2018, Bashivan2016a}.
These maps can then be used to infer the role of specific neurons, or the kind of input they are sensitive to.

Finally, some model inspection techniques were used in a single study.
For instance, in \cite{Ghosh2018}, the class activation map (CAM) technique was extended to overcome its limitations on EEG data.
To use CAMs in a CNN, the channel activations of the last convolutional layer must be averaged spatially before being fed into the model's penultimate layer, which is a \gls{fc} layer.
For a specific input image, a map can then be created to highlight parts of the image that contributed the most to the decision, by computing a weighted average of the last convolutional layer's channel activations.
Other techniques include Deeplift \cite{Lawhern2018}, saliency maps \cite{Vilamala2017}, input-feature unit-output correlation maps \cite{Schirrmeister2017}, retrieval of closest examples \cite{Deiss2018}, analysis of performance with transferred layers \cite{Hajinoroozi2017}, analysis of most-activating input windows \cite{Hartmann2018b}, analysis of generated outputs \cite{Hartmann2018}, and ablation of filters \cite{Lawhern2018}.


\begin{table}[]
\caption{Model inspection techniques used by more than one study.}
\label{tab:model_inspection}
\centering
\begin{tabular}{p{0.5\textwidth}p{0.4\textwidth}}
\toprule
Model inspection technique &                                                                                                                                                                                          Citation \\
\midrule
Analysis of weights                                    &  \cite{Perez-Benitez2018, Yoon2018, Langkvist2018, Deiss2018, Lawhern2018, Xu2016, Tsinalis2016a, Nurse2016, Tabar2016a, Zheng2015, Stober2015, Manor2015, Yang2015a, Langkvist2012, Cecotti2011} \\
Analysis of activations                                &                                                                                           \cite{Yuan2018a, Waytowich2018, Lawhern2018, kwak2017, Yin2017a, Supratak2017, Shamwell2016, Manor2015} \\
Input-perturbation network-prediction correlation maps &                                                                                                              \cite{Schirrmeister2017a, Volker2018, Hartmann2018b, Behncke2017, Schirrmeister2017} \\
Generating input to maximize activation                &                                                                                                                                      \cite{VanPutten2018b, Ruffini2018a, Sors2018, Bashivan2016a} \\
Occlusion of input                                     &                                                                                                                                                        \cite{Lee2018, Chambon2018, Thodoroff2016} \\
\bottomrule
\end{tabular}
\end{table}

\subsection{Reporting of results} \label{sec:reporting}

The performance of DL methods on EEG is of great interest as it is still not clear whether DL can outperform traditional EEG processing pipelines \cite{lotte2018review}.
Thus, a major question we thus aim to answer in this review is:  ``Does DL lead to better performance than traditional methods on EEG?''
However, answering this question is not straightforward, as benchmark datasets, baseline models, performance metrics and reporting methodology all vary considerably between the studies.
In contrast, other application domains of DL, such as computer vision and NLP, benefit from standardized datasets and reporting methodology \cite{goodfellow2016deep}.

Therefore, to provide as satisfying an answer as possible, we adopt a two-pronged approach. First, we review how the studies reported their results by focusing on directly quantifiable items: 1) the type of baseline used as a comparison in each study, 2) the performance metrics, 3) the validation procedure, and 4) the use of statistical testing. Second, based on these points and focusing on studies that reported accuracy comparisons with baseline models, we analyze the reported performance of a majority of the reviewed studies.

\subsubsection{Type of baseline}

When contributing a new model, architecture or methodology to solve an already existing problem, it is necessary to compare the performance of the new model to the performance of state-of-the-art models commonly used for the problem of interest.
Indeed, without a baseline comparison, it is not possible to assess whether the proposed method provides any advantage over the current state-of-the-art.

Points of comparison are typically obtained in two different ways: 1) (re)implementing standard models or 2) referring to published models. 
In the first case, authors will implement their own baseline models, usually using simpler models, and evaluate their performance on the same task and in the same conditions.
Such comparisons are informative, but often do not reflect the actual state of the art on a specific task.
In the second case, authors will instead cite previous literature that reported results on the same task and/or dataset. 
This second option is not always possible, especially when working on private datasets or tasks that have not been explored much in the past.

In the case of typical EEG classification tasks, state-of-the-art approaches usually involve traditional processing pipelines that include feature extraction and shallow/classical machine learning models.
With that in mind, $67.9\%$ of the studies selected included at least one traditional processing pipeline as a baseline model (see Fig.~\ref{fig:reproducibility}).
Some studies instead (or also) compared their performance to DL-based approaches, to highlight incremental improvements obtained by using different architectures or training methodology: $34.0\%$ of the studies therefore included at least one DL-based model as a baseline model.
Out of the studies that did not compare their models to a baseline, six did not focus on the classification of EEG.
Therefore, in total, $21.1\%$ of the studies did not report baseline comparisons, making it impossible to assess the added value of their proposed methods in terms of performance.

\begin{figure}[h]
    \begin{subfigure}[t]{0.5\textwidth}
    \includegraphics[width=\linewidth]{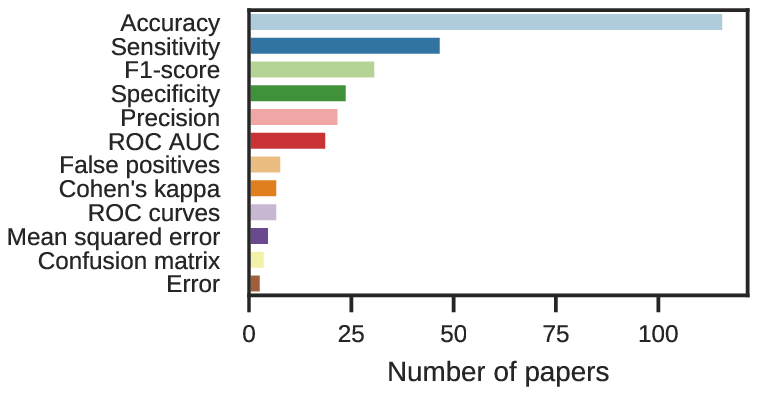}
    \centering
    \caption{Type of performance metrics used in the selected studies. Only metrics that appeared in at least three different studies are included in this figure.}
    \label{fig:performance_metrics}
    \end{subfigure}
    ~
    \begin{subfigure}[t]{0.5\textwidth}
    \includegraphics[width=\linewidth]{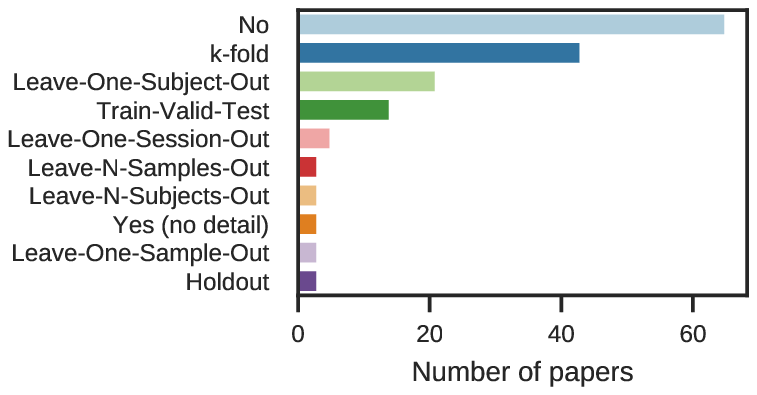}
    \centering
    \caption{Cross-validation approaches.}
    \label{fig:data_cross_validation}
    \end{subfigure}
    
    \caption{Performance metrics and cross-validation approaches.}
    \label{fig:perf}
\end{figure}

\subsubsection{Performance metrics}

The types of performance metrics used by studies focusing on EEG classification are shown in Fig.~\ref{fig:performance_metrics}.
Unsurprisingly, most studies used metrics derived from confusion matrices, such as accuracy, sensitivity, f1-score, \gls{rocauc} and precision.
As highlighted in \cite{Chambon2018, Xu2016}, it is often preferable to use metrics that are robust to class imbalance, such as balanced accuracy, f1-score, and the \gls{rocauc} for binary problems.
This is often the case in sleep or epilepsy recordings, where clinical events are rare.

Studies that did not focus on the classification of EEG signals also mainly used accuracy as a metric. Indeed, these studies generally used a classification task to evaluate model performance, although their main purpose was different (e.g., correcting artifacts). In other cases, performance metrics specific to the study's purpose, such as generating data, were used, e.g., the inception score (\cite{salimans2016improved}), the Fr\'echet inception distance (\cite{heusel2017gans}), as well as custom metrics.

\subsubsection{Validation procedure}

When evaluating a machine learning model, it is important to measure its generalization performance, i.e., how well it performs on unseen data.
In order to do this, it is common practice to divide the available data into a training and a test sets.
When hyperparameters need to be tuned, the performance on the test set cannot be used anymore as an unbiased evaluation of the generalization performance of the model.
Therefore, the training set is divided to obtain a third set called a "validation set" which is used to select the best hyperparameter configuration, leaving the test set to evaluate the performance of the best model in an unbiased way.
However, when the amount of data available is small, dividing the data into different sets and only using a subset for training can seriously undermine the performance of data-hungry models.
A procedure known as "cross-validation" is used in these cases, where the data is broken down into different partitions, which will then successively be used as either training or validation data.

The cross-validation techniques used in the selected studies are shown in Fig.~\ref{fig:data_cross_validation}.
Some studies mentioned using cross-validation but did not provide any details. The category `Train-Valid-Test' includes studies doing random permutations of train/valid, train/test or train/valid/test, as well as studies that mentioned splitting their data into training, validation and test sets but did not provide any details on the validation method. The Leave-One-Out variations correspond to the special case where $N=1$ in the Leave-N-Out versions. $60\%$ of the studies did not use any form of cross-validation. Interestingly, in \cite{Loshchilov2017}, the authors proposed a 'warm restart' within the gradient descent steps to remove the need for a validation set.

\subsubsection{Subject handling} \label{ss:intra_inter_subject}

Whether a study focuses on intra- or inter-subject classification has an impact on the performance. Intra-subject models, which are trained and used on the data of a single subject, often lead to higher performance since the model has less data variability to account for.
However, this means the data the model is trained on is obtained from a single subject, and thus often comprises only a few recordings.
In inter-subject studies, models generally see more data, as multiple subjects are included, but must contend with greater data variability, which introduces different challenges.

In the case of inter-subject classification, the choice of the validation procedure can have a big impact on the reported performance of a model.
The Leave-N-Subject-Out procedure, which uses different subjects for training and for testing, may lead to lower performance, but is applicable to real-life scenarios where a model must be used on a subject for whom no training data is available.
In contrast, using k-fold cross-validation on the combined data from all the subjects often means that the same subjects are seen in both the training and testing sets.
In the selected studies, 22 out of the 108 studies using an inter-subject approach used a Leave-N-Subjects-Out or Leave-One-Subjects-Out procedure.

In the selected studies, $25\%$ focused only on intra-subject classification, $61\%$ focused only on inter-subject classification, $8.3\%$ focused on both, and $4\%$ did not mention it. Obviously, 'N/M' studies necessarily fall under one of the three previous categories. The `N/M' might be due to certain domains using a specific type of experiment (i.e. intra or inter-subject) almost exclusively, thereby obviating the need to mention it explicitly.

\begin{figure}[h]
\includegraphics[width=0.5\textwidth]{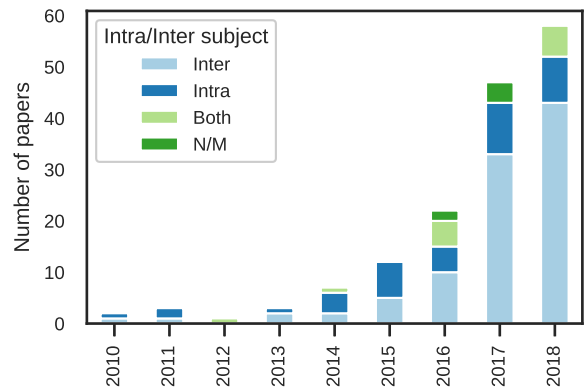}
\centering
\caption{Distribution of intra- vs. inter-subject studies per year.}
\label{fig:data_intra-inter_subject_years}
\end{figure}

Fig.~\ref{fig:data_intra-inter_subject_years} shows that there has been a clear trend over the last few years to leverage DL for inter-subject rather than intra-subject analysis.
In \cite{Deiss2018}, the authors used a large dataset and tested the performance of their model both on new (unseen) subjects and on known (seen) subjects. They obtained $38\%$ accuracy on unseen subjects and $75\%$ on seen subjects, showing that classifying EEG data from unseen subjects can be significantly more challenging than from seen ones.

In \cite{Turner2014}, the authors compared their model on both intra- and inter-subject tasks. Despite the former case providing the model with less less training data than the latter, it led to better results. 
In \cite{Hajinoroozi2016}, the authors compared different DL models and showed that cross-subject (37 subjects) models always performed worse than within-subject models. In \cite{Page2016}, a hybrid system trained on multiple subjects and then fine-tuned on subject-specific data led to the best performance. Finally, in \cite{Thodoroff2016}, the authors compared their DNN to a state-of-the-art traditional approach and showed that deep networks generalize better, although their performance on intra-subject classification is still higher than on inter-subject classification.

\subsubsection{Statistical testing}

To assess whether a proposed model is actually better than a baseline model, it is useful to use statistical tests.
In total, $19.9\%$ of the selected studies used statistical tests to compare the performance of their models to baseline models.
The tests most often used were Wilcoxon signed-rank tests, followed by ANOVAs.

\subsubsection{Comparison of results} \label{ss:comparison_results}

Although, as explained above, many factors make this kind of comparison imprecise, we show in this section how the proposed approaches and traditional baseline models compared, as reported by the selected studies.

We focus on a specific subset of the studies to make the comparison more meaningful.
First, we focus on studies that report accuracy as a direct measure of task performance.
As shown in Fig.~\ref{fig:performance_metrics}, this includes the vast majority of the studies.
Second, we only report studies which compared their models to a traditional baseline, as we are interested in whether DL leads to better results than non-DL approaches.
This means studies which only compared their results to other DL approaches are not included in this comparison.
Third, some studies evaluated their approach on more than one task or dataset.
In this case, we report the results on the task that has the most associated baselines.
If that is more than one, we either report all tasks, or aggregate them if they are very similar (e.g., binary classification of multiple mental tasks, where performance is reported for each possible pair of tasks).
In the case of multimodal studies, we only report the performance on the EEG-only task, if it is available.
Finally, when reporting accuracy differences, we focus on the difference between the best proposed model and the best baseline model, per task.
Following these constraints, a total of $102$ studies/tasks were left for our analysis.

\begin{figure}
\includegraphics[width=\linewidth]{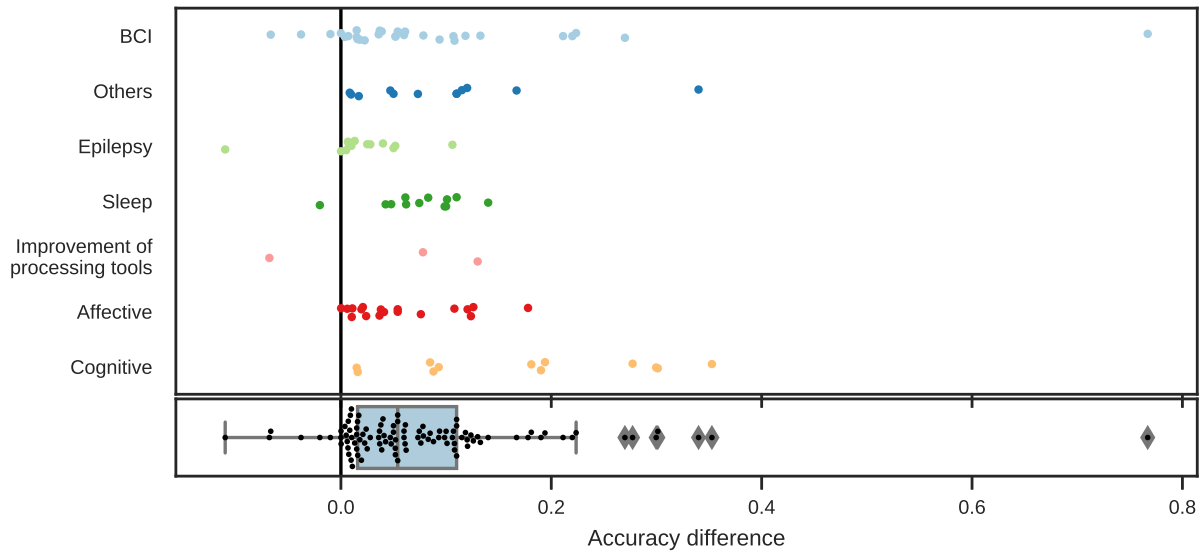}
\centering
\caption{Difference in accuracy between each proposed DL model and corresponding baseline model for studies reporting accuracy (see Section~\ref{ss:comparison_results} for a description of the inclusion criteria). The difference in accuracy is defined as the difference between the best DL model and the best corresponding baseline. In the top figure, each study/task is represented by a single point, and studies are grouped according to their respective domains. The bottom figure is a box plot representing the overall distribution.}
\label{fig:reported_accuracy}
\end{figure}

Figure~\ref{fig:reported_accuracy} shows the difference in accuracy between each proposed model and corresponding baseline per domain type (as categorized in Fig.~\ref{fig:domaintree}), as well as the corresponding distribution over all included studies and tasks.

The median gain in accuracy with DL is of $5.4\%$, with an interquartile range of $9.4\%$.
Only four values were negative values, meaning the proposed DL approach led to a lower performance than the baseline.
The best improvement in accuracy was obtained by \cite{Spampinato2017}, where their approach led to a gain of $76.7\%$ in accuracy in an \gls{rsvp} classification task.

\subsection{Reproducibility} \label{sec:reproducibility}

Reproducibility is a cornerstone of science~\cite{Ioannidis2017}: having reproducible results is fundamental to moving a field forward, especially in a field like machine learning where new ideas spread very quickly. Here, we evaluate ease with which the results of the selected papers can be reproduced by the community using two key criteria: the availability of their data and the availability of their code.

\begin{figure}[h]
\includegraphics[width=\textwidth]{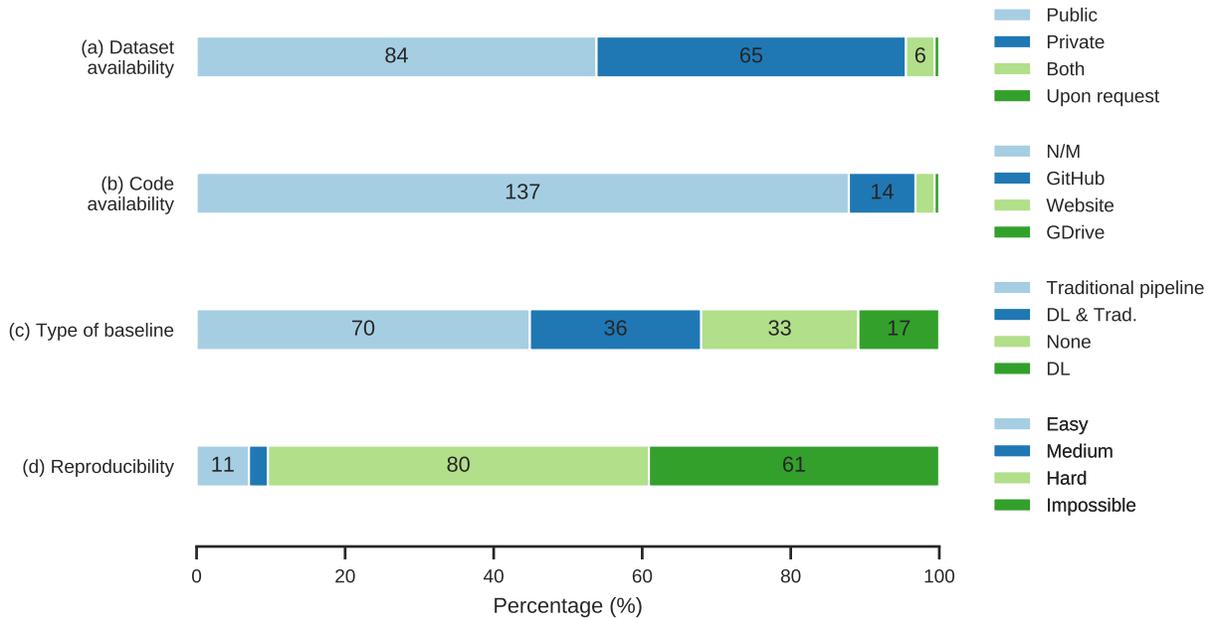}
\centering
\caption{Reproductibility of the selected studies. (a) Availability of the datasets used in the studies, (b) availability of the code, shown by where the code is hosted, (c) type of baseline used to evaluate the performance of the trained models and (d) estimated reproducibility level of the studies (Easy: both the data and the code are available, Medium: the code is available but some data is not publicly available, Hard: either the code or the data is available but not both, Impossible: neither the data nor the code are available).}
\label{fig:reproducibility}
\end{figure}

From the 156 studies reviewed, $54\%$ used public data, $42\%$ used private data\footnote{Data that is not freely available online was considered private regardless of when and where it was recorded. Moreover, three of the reviewed studies mentioned that their data was available upon request but were included in the "private" category.}, and $4\%$ used both public and private data.
In particular, studies focusing on BCI, epilepsy, sleep and affective monitoring made use of openly available datasets the most (see Table~\ref{tab:Datasets}). Interestingly, in cognitive monitoring, no publicly available datasets were used, and papers in that field all relied on internal recordings. 

Fittingly, a total of 33 papers (21\%) explicitly mentioned that more publicly available data is required to support research on DL-EEG. In clinical settings, the lack of labeled data, rather than the quantity of data, was specifically pointed out as an obstacle.

As for the source code, only $19\%$ of the studies chose to make it available online \cite{Kuanar2018, Schirrmeister2017a, Sors2018, Zhang2018a, Wu2018, Schwabedal2018, Lawhern2018, Schirrmeister2017, Zhang2017g, Zhang2017d, Supratak2017, Zhang2017e, Zhang2017c, Loshchilov2017, Spampinato2017, Bashivan2016a, Stober2015, Stober2014, Langkvist2012} and as illustrated in Fig~\ref{fig:reproducibility}, GitHub is by far the preferred code sharing platform. Needless to say, having access to the source code behind published results can drastically reduce time and increase incentive to reproduce a paper's results.

Therefore, taking both data and code availability into account, only 11 out of 156 studies ($7\%$) could easily be reproduced using both the same data and code \cite{Schirrmeister2017a, Sors2018, Schwabedal2018, Zhang2017g, Zhang2017d, Supratak2017, Zhang2017c, Loshchilov2017, Spampinato2017, Stober2015, Langkvist2012}. 4 out of 156 studies ($3\%$) shared their code but tested on both private and public data making their studies only partially reproducible \cite{Zhang2018a, Lawhern2018, Schirrmeister2017, Zhang2017e}, see Fig.~\ref{fig:reproducibility}. 
As follows, a significant number of studies (61) did not have publicly available data or code, making them almost impossible to reproduce.

It is important to note, moreover, that for the results of a study to be perfectly reproduced, the authors would also need to share the weights (i.e.  parameters) of the network.  Sharing the code and the architecture of the network might not be sufficient since retraining the network could converge to a different minimum. On the other hand, retraining the network could also end up producing better results if a better performing model is obtained. For recommendations on how to best share the results, the code, the data and relevant information to make a study easy to reproduce, please see the discussion section and the checklist provided in Appendix~\ref{appendix:check_list}.

\begin{table}
\caption{Most often used datasets by domain. Datasets that were only used by one study are grouped under "Other" for each category.}
\begin{tabular}{llrp{5cm}}
\toprule
Main domain & Dataset &  \# articles & References \\
\midrule
Affective & DEAP &            9 &                                                            \cite{Li2018, Alhagry2017, BenSaid2017a, Lin2017, Xu2016, Liu2016, Frydenlund2015, Jirayucharoensak2014, Li2013} \\
      & SEED &            3 &                                                                                                                                        \cite{Zhang2018, Liu2016, Zheng2015} \\
\hline
BCI & BCI Competition &           13 &  \cite{Gao2018, Lawhern2018, Sakhavi2017, Schirrmeister2017, Schirrmeister2017, Tabar2016a, Tabar2016a, Manor2015, Sakhavi2015, Yang2015a, Ding2015, Ding2015, Cecotti2011} \\
      & Other &            8 &                                                        \cite{Hasib2018, Lawhern2018, Schirrmeister2017, Hajinoroozi2017, Hajinoroozi2017, Hajinoroozi2017, Sun2016, An2016} \\
      & eegmmidb &            8 &                                                                    \cite{Zhang2018c, Major2017, Zhang2017g, Zhang2017d, Zhang2017a, Dharamsi2017, Normand2015, Alomari2013} \\
      & Keirn \& Aunon (1989) &            2 &                                                                                                                                           \cite{Padmanabh2017, Patnaik2017} \\
      & MAHNOB &            1 &                                                                                                                                                      \cite{drouin2016using} \\
\hline
Cognitive & Other &            4 &                                                                                                        \cite{Kuanar2018, Hajinoroozi2015, Hajinoroozi2015, Hajinoroozi2015} \\
      & EEG Eye State &            1 &                                                                                                                                                           \cite{Narejo2016} \\
\hline
Epilepsy & CHB-MIT &            9 &                                                           \cite{Yan2018, Tsiouris2018, Yuan2018a, Truong2018, Truong2018a, Page2016, Thodoroff2016, Pramod2015, Turner2014} \\
      & Bonn University &            7 &                                                                 \cite{Hussein2018, Ullah2018, Ahmedt-Aristizabal2018, Talathi2017, Acharya2017, Omerhodzic2013, Naderi2010} \\
      & TUH &            5 &                                                                                          \cite{Golmohammadi2017b, Shah2017, Golmohammadi2017a, Golmohammadi2017, Yang2016b} \\
      & Other &            3 &                                                                                                                              \cite{Truong2018, Golmohammadi2017a, Taqi2017} \\
      & Freiburg Hospital &            2 &                                                                                                                                              \cite{Truong2018, Truong2018a} \\
\hline
Generation of data & BCI Competition &            2 &                                                                                                                                               \cite{Corley2018, Zhang2018b} \\
      & MAHNOB &            1 &                                                                                                                                                             \cite{Wang2018} \\
      & Other &            1 &                                                                                                                                                       \cite{Schwabedal2018} \\
      & SEED &            1 &                                                                                                                                                             \cite{Wang2018} \\
\hline
Improvement of processing tools & BCI Competition &            3 &                                                                                                                                       \cite{Yang2018, Sturm2016, Yang2016a} \\
      & Other &            2 &                                                                                                                                                \cite{Yepes2017, Stober2015} \\
      & Bonn University &            1 &                                                                                                                                                              \cite{Wen2018} \\
      & CHB-MIT &            1 &                                                                                                                                                              \cite{Wen2018} \\
\hline
Others & TUH &            3 &                                                                                                                              \cite{Schirrmeister2017a, Roy2018, Zhang2018a} \\
      & eegmmidb &            3 &                                                                                                                                   \cite{Zhang2018a, Zhang2017e, Zhang2017c} \\
      & Other &            2 &                                                                                                                                           \cite{VanPutten2018b, Zhang2018a} \\
      & EEG Eye State &            1 &                                                                                                                                                             \cite{Lee2018a} \\
\hline
Sleep & MASS &            4 &                                                                                                                   \cite{Phan2018, Chambon2018, Supratak2017, dong2018mixed} \\
      & Sleep EDF &            4 &                                                                                                                   \cite{Vilamala2017, Supratak2017, Xie2017, Tsinalis2016a} \\
      & Other &            3 &                                                                                                                                     \cite{Sors2018, Tripathy2018, Giri2016} \\
      & UCDDB &            3 &                                                                                                                           \cite{Langkvist2018, Manzano2017a, Langkvist2012} \\
\bottomrule
\end{tabular}
\label{tab:Datasets}
\end{table}




\section{Discussion} \label{sec:discussion}

In this section, we review the most important findings from our results section, and discuss the significance and impact of various trends highlighted above. We also provide recommendations for DL-EEG studies and present a checklist to ensure reproducibility in the field.

\subsection{Rationale}

It was expected that most papers selected for the review would focus on the classification of EEG data, as DL has historically led to important improvements on supervised classification problems \cite{lecun2015deep}.
Interestingly though, several papers also focused on new applications that were made possible or facilitated by DL: for instance, generating images conditioned on EEG, generating EEG, transfer learning between subjects, or feature learning. One of the main motivations for using DL cited by the papers reviewed was the ability to use raw EEG with no manual feature extraction steps.
We expect these kinds of applications that go beyond using DL as a replacement for traditional processing pipelines to gain in popularity.


\subsection{Data}

A critical question concerning the use of DL with EEG data remains ``How much data is enough data?''.
In Section~\ref{sec:data}, we explored this question by looking at various descriptive dimensions: the number of subjects, the amount of EEG recorded, the number of training/test/validation examples, the sampling rate and data augmentation schemes used.

Although a definitive answer cannot be reached, the results of our meta-analysis show that the amount of data necessary to at least match the performance of traditional approaches is already available. Out of the 156 papers reviewed, only six reported lower performance for DL methods over traditional benchmarks. To achieve these results with limited amounts of data, shallower architectures were often preferred. Data augmentation techniques were also used successfully to improve performance when only limited data was available. However, more work is required to clearly assess their advantages and disadvantages. Indeed, although many studies used overlapping sliding windows, there seems to be no consensus on the best overlapping percentage to use, e.g., the impact of using a sliding window with 1\% overlap versus 95\% overlap is still not clear. \Gls{bci} studies had the highest variability for this hyperparameter, while clinical applications such as sleep staging already appeared more standardized with most studies using 30\,s non-overlapping windows.

Many authors concluded their paper suggesting that having access to more data would most likely improve the performance of their models. With large datasets becoming public, such as the TUH Dataset \cite{harati2014tuh} and the National Sleep Research Resource \cite{zhang2018national}, deeper architectures similar to the ones used in computer vision might become increasingly usable. However, it is important to note that the availability of data is quite different across domains. In clinical fields such as sleep and epilepsy, data usually comes from hospital databases containing years of recordings from several patients, while other fields usually rely on data coming from lab experiments with a limited number of subjects.

The potential of DL in EEG also lies in its ability (at least in theory) to generalize across subjects and to enable transfer learning across tasks and domains. When only limited data is available, intra-subject models still work best given the inherent subject variability of EEG data. However, transfer learning might be the key to moving past this limitation. Indeed, Page and colleagues \cite{Page2016} showed that with hybrid models, one can train a neural network on a pool of subjects and then fine-tune it on a specific subject, achieving good performances without needing as much data from a specific subject.

While we did report the sampling rate, we did not investigate its effect on performance because no relationship stood out particularly in any of the reviewed papers. The impact of the number of channels though, was specifically studied. For example, in \cite{Chambon2018}, the authors showed that they could achieve comparable results with a lower number of channels. As shown in Fig.~\ref{fig:data_eeg_hardware}, a few studies used low-cost EEG devices, typically limited to a lower number of channels. These more accessible devices might therefore benefit from DL methods, but could also enable faster data collection on a larger-scale, thus facilitating DL in return.

As DL-EEG is highly data-driven, it is important when publishing results to clearly specify the amount of data used and to clarify terminology (see Table~\ref{tab:terminology} for an example). We noticed that many studies reviewed did not clearly describe the EEG data that they used (e.g., the number of subjects, number of sessions, window length to segment the EEG data, etc.) and therefore made it hard or impossible for the reader to evaluate the work and compare it to others. Moreover, reporting learning curves (i.e. performance as a function of the number of examples) would give the reader valuable insights on the bias and variance of the model. 

\subsection{EEG processing}
According to our findings, the great majority of the reviewed papers preprocessed the EEG data before feeding it to the deep neural network or extracting features. Despite observing this trend, we also noticed that recent studies outperformed their respective baseline(s) using completely raw EEG data. Almogbel et al. \cite{Almogbel2018} used raw EEG data to classify cognitive workload in vehicle drivers, and their best model achieved a classification accuracy approximately $4\%$ better than their benchmarks which employed preprocessing on the EEG data. Similarly, Aznan et al. \cite{Aznan2018} outperformed the baselines by a $4\%$ margin on SSVEP decoding using no preprocessing. Thus, the answer to whether it is necessary to preprocess EEG data when using DNNs remains elusive.

As most of the works considered did not use, or explicitly mention using, artifact removal methods, it appears that this EEG processing pipeline step is in general not required. However, one should observe that in specific cases such as tasks that inherently elicit quick eye movements (MATB-II \cite{comstock1992multi}), artifact handling might still be crucial to obtaining desired performance.    

One important aspect we focused on is whether it is necessary to use EEG features as inputs to \glspl{dnn}. After analyzing the type of input used by each paper, we observed that there was no clear preference for using features or raw EEG time-series as input. We noticed though that most of the papers using CNNs used raw EEG as input. With \glspl{cnn}  becoming increasingly popular, one can conclude that there is a trend towards using raw EEG instead of hand-engineered features. This is not surprising, as we observed that one of the main motivations mentioned for using \glspl{dnn} on EEG processing is to automatically learn features.
Furthermore, frequency-based features, which are widely used as hand-crafted features in EEG \cite{lotte2018review}, are very similar to the temporal filters learned by a \gls{cnn}. Indeed, these features are often extracted using Fourier filters which apply a convolutive operation. This is also the case for the temporal filters learned by a \gls{cnn} although in the case of \glspl{cnn} the filters are learned. 

From our analysis, we also aimed to identify which input type should be used when trying to solve a problem from scratch. While the answer depends on many factors such as the domain of application, we observed that in some cases raw EEG as input consistently outperformed baselines based using classically extracted features. For example, for seizure classification, recently proposed models using raw EEG data as input \cite{Hao2018,Ullah2018,Shea2018} achieved better performances than classical baseline methods, such as \glspl{svm} with frequency-domain features. For this particular task, we believe following the current trend of using raw EEG data is the best way to start exploring a new approach. 


\subsection{Deep learning methodology}

Another major topic this review aimed at covering is the \gls{dl} methodology itself.
Our analysis focused on architecture trends and training decisions, as well as on model selection techniques.

\subsubsection{Architecture}

Given the inherent temporal structure of EEG, we expected RNNs would be more widely employed than models that do not explicitly take time dependencies into account. 
However, almost half of the selected papers used CNNs. This observation is in line with recent discussions and findings regarding the effectiveness of CNNs for processing time series \cite{bai2018empirical}. We also noticed that the use of energy-based models such as RBMs has been decreasing, whereas on the other hand, popular architectures in the computer vision community such as GANs have started to be applied to EEG data as well.

Moreover, regarding architecture depth, most of the papers used fewer than five layers. When comparing this number with popular object recognition models such as VGG and ResNet for the ImageNet challenge comprising 19 and 34 layers respectively, we conclude that for EEG data, shallower networks are currently necessary. Schirrmeister et al. \cite{schirrmeister2017deep} specifically focused on this aspect, comparing the performance of architectures with different depths and structures, such as fully convolutional layers and residual blocks, on different tasks. Their results showed that in most cases, shallower fully convolutional models outperformed their deeper counterpart and architectures with residual connections.

\subsubsection{Training and optimization}

Although crucial to achieving good results when using neural networks, only $21\%$ of the papers employed some hyperparameter search strategy. Even fewer studies provided detailed information about the method used. Amongst these, Stober et al. \cite{Stober2015} described their hyperparameter selection method and cited its corresponding implementation; in addition, the available budget in number of iterations per searching trial as well as the cross-validation split were mentioned in the paper.

\subsubsection{Model inspection}

Inspecting trained \gls{dl} models is important, as DNNs are notoriously seen as black boxes, when compared to more traditional methods.
This is problematic in clinical settings for instance, where understanding and explaining the choice made by a classification model might be critical to making informed clinical choices.
Neuroscientists might also be interested by what drives a model's decisions and use that information to shape hypotheses about brain function.

About $27\%$ of the reviewed papers looked at interpreting their models.
Interesting work on the topic, specifically tailored to EEG, was reviewed in \cite{Schirrmeister2017, Hartmann2018b, Ghosh2018}.
Sustained efforts aimed at inspecting models and understanding the patterns they rely on to reach decisions are necessary to broaden the use of DL for EEG processing.

\subsection{Reported results}

Our meta-analysis focused on how studies compared classification accuracy between their models and traditional EEG processing pipelines on the same data.
Although a great majority of studies reported improvements over traditional pipelines, this result has to be taken with a grain of salt.
First, the difference in accuracy does not tell the whole story, as an improvement of $10\%$, for example, is typically more difficult to achieve from $80$ to $90\%$ than from $40$ to $50\%$.
More importantly though, very few articles reported negative improvements, which could be explained by a publication bias towards positive results.

The reported baseline comparisons were highly variable: some used simple models (e.g., combining straightforward spectral features and linear classifiers), others used more sophisticated pipelines (including multiple features and non-linear approaches), while a few reimplemented or cited state-of-the-art models that were published on the same dataset and/or task.
Since the observed improvement will likely be higher when comparing to simple baselines than to state-of-the-art results, the values that we report might be biased positively.
For instance, only two studies used Riemannian geometry-based processing pipelines as baseline models \cite{Aznan2018, Lawhern2018}, although these methods have set a new state-of-the-art in multiple EEG classification tasks \cite{lotte2018review}.

Moreover, many different tasks and thus datasets were used.
These datasets are often private, meaning there is very limited or no previous literature reporting results on them.
On top of this, the lack of reproducibility standards can lead to low accountability: since study results are not expected to be replicated and results can be inflated by non-standard practices such as omitting cross-validation.

Different approaches have been taken to solve the problem of heterogeneity of result reporting and benchmarking in the field of machine learning.
For instance, OpenML \cite{OpenML2013} is an online platform that facilitates the sharing and running of experiments, as well as the benchmarking of models.
As of November 2018, the platform already contained one EEG dataset and multiple submissions.
The MOABB \cite{jayaram2018moabb}, a solution tailored to the field of brain-computer interfacing, is a software framework for ensuring the reproducibility of BCI experiments and providing public benchmarks for many BCI datasets.
In \cite{heilmeyer2018large}, a similar approach, but for DL specifically, is proposed.

Additionally, a few EEG/MEG/ECoG classification online competitions have been organized in the last years, for instance on the Kaggle platform (see Table~1 of \cite{congedo2017riemannian}).
These competitions informally act as benchmarks: they provide a standardized dataset with training and test splits, as well as a leaderboard listing the performance achieved by every competitor.
These platforms can then be used to evaluate the state-of-the-art as they provide a publicly available comparison point for new proposed architectures.
For instance, the IEEE NER 2015 Conference competition on error potential decoding could have been used as a benchmark for the studies reviewed that focused on this topic.

Making use of these tools, or extending them to other EEG-specific tasks, appears to be one of the greatest challenges for the field of DL-EEG at the moment, and might be the key to more efficient and productive development of practical EEG applications.
Whenever possible, authors should make sure to provide as much information as possible on the baseline models they have used, and explain how to replicate their results (see Section~\ref{ssec:reproducibility_discussion}).

\subsection{Reproducibility} \label{ssec:reproducibility_discussion}

The significant use of public EEG datasets across the reviewed studies suggests that open data has greatly contributed to recent developments in DL-EEG. On the other hand, $42\%$ of studies used data not publicly available - notably in domains such as cognitive monitoring. To move the field forward, it is thus important to create new benchmark datasets and share internal recordings. Moreover, the great majority of papers did not make their code available. Many papers reviewed are thus more difficult to reproduce: the data is not available, the code has not been shared, and the baseline models that were used to compare the performances of the models are either non-existent or not available.

Recent initiatives to promote best practices in data and code sharing would benefit the field of DL-EEG. 
FAIR neuroscience \cite{wilkinson2016fair} and the Brain Imaging Data Structure (BIDS) \cite{gorgolewski2016brain} both provide guidelines and standards on how to acquire, organize and share data and code. 
BIDS extensions specific to EEG \cite{pernet2018bidseeg} and \gls{meg} \cite{niso2018meg} were also recently proposed.
Moreover, open source software toolboxes are available to perform DL experiments on EEG. For example, the recent toolbox developed by Schirrmeister and colleagues, called BrainDecode \cite{Schirrmeister2017}, enables faster and easier development cycles by providing the basic functionality required for DL-EEG analysis while offering high level and easy to use functions to the user. 
The use of common software tools could facilitate reproducibility in the community.
Beyond reproducibility, we believe simplifying access to data, making domain knowledge accessible and sharing code will enable more people to jump into the field of DL-EEG and contribute, transforming what has traditionally been a domain-specific problem into a more general problem that can be tackled with machine learning and DL methods.

\subsection{Recommendations}

To improve the quality and reproducibility of the work in the field of DL-EEG, we propose six guidelines in Table~\ref{tab:recommendations}.
Moreover, Appendix~\ref{appendix:check_list} presents a checklist of items that are critical to ensuring reproducibility and should be included in future studies.

\begin{table}[h]
\caption{Recommendations for future DL-EEG studies. See Appendix~\ref{appendix:check_list} for a detailed list of items to include.}
\label{tab:recommendations}
\centering
\begin{tabular}{cp{0.33\textwidth}p{0.55\textwidth}}
\hline
  & Recommendation & Description \\ \hline
1 & \textbf{Clearly describe the architecture.} & Provide a table or figure clearly describing your model (e.g., see \cite{Chambon2018, Golmohammadi2017b, Schirrmeister2017}).\\
2 & \textbf{Clearly describe the data used.} & Make sure the number of subjects, the number of examples, the data augmentation scheme, etc. are clearly described. Use unambiguous terminology or define the terms used (for an example, see Table~\ref{tab:terminology}). \\
3 & \textbf{Use existing datasets.} & Whenever possible, compare model performance on public datasets.\\
4 & \textbf{Include state-of-the-art baselines.} & If focusing on a research question that has already been studied with traditional machine learning, clarify the improvements brought by using DL.\\
5 & \textbf{Share internal recordings.} & Whenever possible.\\
6 & \textbf{Share reproducible code.} & Share code (including hyperparameter choices and model weights) that can easily be run on another computer, and potentially reused on new data.\\
\hline
\end{tabular}
\end{table}



\subsubsection{Supplementary material}

Along with the current paper, we make our data items table and related code available online at \url{http://dl-eeg.com}.
We encourage interested readers to consult it in order to dive deeper into data items that are of specific interest to them - it should be straightforward to reproduce and extend the results and figures presented in this review using the code provided.
The data item table is intended to be updated frequently with new articles, therefore results will be brought up to date periodically.

Authors of DL-EEG papers not included in the review are invited to submit a summary of their article following the format of our data items table to our online code repository.
We also invite authors whose papers are already included in the review to verify the accuracy of our summary.
Eventually, we would like to indicate which studies have been submitted or verified by the original authors.

By updating the data items table regularly and inviting researchers in the community to contribute, we hope to keep the supplementary material of the review relevant and up-to-date as long as possible.

\subsection{Limitations}

In this section, we quickly highlight some limitations of the present work.
First, our decision to include arXiv preprints in the database search requires some justification. 
It is important to note that arXiv papers are not peer-reviewed. 
Therefore, some of the studies we selected from arXiv might not be of the same quality and scientific rigor as the ones coming from peer-reviewed journals or conferences. 
For this reason, whenever a preprint was followed by a publication in a peer-reviewed venue, we focused our analysis on the peer-reviewed version.
ArXiv has been largely adopted by the DL community as a means to quickly disseminate results and encourage fast research iteration cycles.
Since the field of DL-EEG is still young and a limited number of publications was available at the time of writing, we decided to include all the papers we could find, knowing that some of the newer trends would be mostly visible in repositories such as arXiv. 
Our goal with this review was to provide a transparent and objective analysis of the trends in DL-EEG.
By including preprints, we feel we provided a better view of the current state-of-the-art, and are also in a better position to give recommendations on how to share results of DL-EEG studies moving forward.

Second, in order to keep this review reasonable in length, we decided to focus our analysis on the points that we judged most interesting and valuable.
As a result, various factors that impact the performance of DL-EEG were not covered in the review.
For example, we did not cover weight initialization: in \cite{Golmohammadi2017b}, the authors compared 10 different initialization methods and showed an impact on the specificity metric, with ranged from $85.1\%$ to $96.9\%$.
Similarly, multiple data items were collected during the review process, but were not included in the analysis.
These items, which include data normalization procedures, software toolboxes, hyperparameter values, loss functions, training hardware, training time, etc., remain available online for the interested reader.
We are confident other reviews or research articles will be able to focus on more specific elements.

Third, as any literature review in a field that is quickly evolving, the relevance of our analysis decays with time as new articles are being published and new trends are established.
Since our last database search, we have already identified other articles that should eventually be added to the analysis.
Again, making this work a living review by providing the data and code online will hopefully ensure the review will be of value and remain relevant for years to come.


\section{Conclusion} 
\label{sec:conclusion}

The usefulness of EEG as a functional neuroimaging tool is unequivocal: clinical diagnosis of sleep disorders and epilepsy, monitoring of cognitive and affective states, as well as brain-computer interfacing all rely heavily on the analysis of EEG.
However, various challenges remain to be solved.
For instance, time-consuming tasks currently carried out by human experts, such as sleep staging, could be automated to increase the availability and flexibility of EEG-based diagnosis.
Additionally, better generalization performance between subjects will be necessary to truly make \glspl{bci} useful.
DL has been proposed as a potential candidate to tackle these challenges.
Consequently, the number of publications applying DL to EEG processing has seen an exponential increase over the last few years, clearly reflecting a growing interest in the community in these kinds of techniques.

In this review, we highlighted current trends in the field of DL-EEG by analyzing 156 studies published between January 2010 and July 2018 applying DL to EEG data. 
We focused on several key aspects of the studies, including their origin, rationale, the data they used, their EEG processing methodology, DL methodology, reported results and level of reproducibility.

Among the major trends that emerged from our analysis, we found that 1) DL was mainly used for classifying EEG in domains such as brain-computer interfacing, sleep, epilepsy, cognitive and affective monitoring, 2) the quantity of data used varied a lot, with datasets ranging from 1 to over 16,000 subjects (mean = 223; median = 13), producing to 62 up to 9,750,000 examples (mean = 251,532; median = 14,000) and from two to 4,800,000 minutes of EEG recording (mean = 62,602; median = 360), 3) various architectures have been used successfully on EEG data, with \glspl{cnn}, followed by \glspl{rnn} and \glspl{ae}, being most often used, 4) there is a clear growing interest towards using raw EEG as input as opposed to handcrafted features, 5) almost all studies reported a small improvement from using DL when compared to other baselines and benchmarks (median = $5.4\%$), and 6) while several studies used publicly available data, only a handful shared their code - the great majority of studies reviewed thus cannot easily be reproduced.


Moreover, given the high variability in how results were reported, we made six recommendations to ensure reproducibility and fair comparison of results: 1) clearly describe the architecture, 2) clearly describe the data used, 3) use existing datasets, whenever possible, 4) include state-of-the-art baselines, ideally using the original authors' code, 5) share internal recordings, whenever possible, and 6) share code, as it is the best way to allow others to pick up where your work leaves off.
We also provided a checklist (see Appendix~\ref{appendix:check_list}) to help authors of DL-EEG studies make sure all the relevant information is available in their publications to allow straightforward reproduction.

Finally, to help the DL-EEG community maintain an up-to-date list of published work, we made our data items table open and available online. 
The code to reproduce the statistics and figures of this review as well as the full summaries of the papers are also available at \url{http://dl-eeg.com}.

The current general interest in artificial intelligence and DL has greatly benefited various fields of science and technology. 
Advancements in other field of application will most likely benefit the neuroscience and neuroimaging communities in the near future, and enable more pervasive and powerful applications based on EEG processing.
We hope this review will constitute a good entry point for EEG researchers interested in applying DL to their data, as well as a good summary of the current state of the field for DL researchers looking to apply their knowledge to new types of data.


\section*{Acknowledgments}
We thank Raymundo Cassani, Colleen Gillon, Jo\~ao Monteiro and William Thong for comments that greatly improved the manuscript.

\section*{Funding}
This work was supported by the Natural Sciences and Engineering Research Council of Canada (NSERC) for YR (reference number: RDPJ 514052-17), HB, IA and THF, the Fonds qu\'eb\'ecois de la recherche sur la nature et les technologies (FRQNT) for YR and InteraXon Inc. (graduate funding support) for HB.

\clearpage

\bibliographystyle{acm}
\bibliography{ms.bib}

\begin{appendices}

\section{List of acronyms}
\printglossary[type=\acronymtype,title={},nogroupskip]
\glsaddallunused

\section{Checklist of items to include in a DL-EEG study} \label{appendix:check_list}

\newlist{todolist}{itemize}{2}
\setlist[todolist]{label=$\square$}

This section contains a checklist of items we believe DL-EEG papers should mention to ensure their published results are readily reproducible.
The following items of information should all be clearly stated at one point or another in the text or supplementary materials of future DL-EEG studies:

\paragraph{Data}
\begin{todolist}
    \item Number of subjects (and relevant demographic data)
    \item Electrode montage including reference(s) (number of channels and their locations)
    \item Shape of one example (e.g., ``$256$ samples $\times$ $16$ channels'')
    \item Data augmentation technique (e.g., percentage of overlap for sliding windows)
    \item Number of examples in training, validation and test sets
\end{todolist}
\paragraph{EEG processing}
\begin{todolist}
  \item Temporal filtering, if any
  \item Spatial filtering, if any
  \item Artifact handling techniques, if any
  \item Resampling, if any
\end{todolist}
\paragraph{Neural network architecture}
\begin{todolist}
    \item Architecture type
    \item Number of layers (consider including a diagram or table to represent the architecture)
    \item Number of learnable parameters
\end{todolist}
\paragraph{Training hyperparameters}
\begin{todolist}
    \item Parameter initialization
    \item Loss function
    \item Batch size
    \item Number of epochs
    \item Stopping criterion
    \item Regularization (e.g., dropout, weight decay, etc.)
    \item Optimization algorithm (e.g., stochastic gradient descent, Adam, RMSProp, etc.)
    \item Learning rate schedule and optimizer parameters
    \item Values of \textbf{all} hyperparameters (including random seed) for the results that are presented in the paper
    \item Hyperparameter search method
\end{todolist}
\paragraph{Performance and model comparison}
\begin{todolist}
    \item Performance metrics (e.g., f1-score, accuracy, etc.)
    \item Type of validation scheme (intra- vs. inter-subject, leave-one-subject-out, k-fold cross-validation, etc.)
    \item Description of baseline models (thorough description or reference to published work)
\end{todolist}


\end{appendices}

\end{document}